\documentclass[a4paper,fleqn]{cas-sc}
\usepackage{subcaption} % cho vào phần 
\usepackage{algorithmic}
\usepackage{algorithm}

\usepackage[authoryear,longnamesfirst]{natbib}
\usepackage{tikz, pgfplots}
\pgfplotsset{compat=1.18}
\def\tsc#1{\csdef{#1}{\textsc{\lowercase{#1}}\xspace}}
\tsc{WGM}
\tsc{QE}

\begin{document}
\let\WriteBookmarks\relax
\def\floatpagepagefraction{1}
\def\textpagefraction{.001}

\shorttitle{}    

\shortauthors{}

\shorttitle{}    

\shortauthors{}  

\title [mode = title]{Energy-Efficient IoT Application for Real-Time Multi-Student Emotion Recognition to Improve Classroom Interaction}  

\tnotemark[1] 

\tnotetext[1]{} 

\author[1]{Hai Nguyen Nam}[type=editor,
                        auid=000,
                        orcid=0009-0009-4445-2811]
\cormark[1]
\fnmark[1]
\ead{HieuDT.b20cn242@stu.ptit.edu.vn}
\ead[url]{https://orcid.org/0009-0009-4445-2811}

\author[1]{Hieu Dao Trong}[
                        orcid=0009-0006-0623-4539]
\fnmark[2]
\ead{hainn.b21cn320@stu.ptit.edu.vn}
\ead[url]{https://orcid.org/0009-0006-0623-4539}

\author[1]{Nam Vu}[]
\fnmark[3]
\ead{nvhung_vt1@ptit.edu.vn}

\author[1]{Hung Nguyen Viet}[]
\fnmark[4]
\ead{nvhung_vt1@ptit.edu.vn}

\author[1]{Cong Tran}[
                        orcid=0000-0001-9467-4978]
\fnmark[5]
\ead{congtt@ptit.edu.vn}
\ead[url]{https://orcid.org/0000-0001-9467-4978}

\affiliation[1]{organization={Posts and Telecommunications Institute of Technology (PTIT)},
                addressline={Km 10 Nguyen Trai Street, Ha Dong District},
                city={Hanoi},
                country={Vietnam}}

\cortext[cor1]{Corresponding author}

\begin{abstract}
This study presents high-throughput, real-time multi-agent affective computing framework designed to enhance classroom learning through emotional state monitoring. As large classroom sizes and limited teacher–student interaction increasingly challenge educators, there is a growing need for scalable, data-driven tools capable of capturing students’ emotional and engagement patterns in real time. The system was evaluated using the Classroom Emotion Dataset, consisting of 1,500 labeled images and 300 classroom detection videos. Tailored for IoT devices, the system addresses load balancing and latency challenges through efficient real-time processing. Field testing was conducted across three educational institutions in a large metropolitan area: a primary school (hereafter school A), a secondary school (school B), and a high school (school C). The system demonstrated robust performance, detecting up to 50 faces at 25 FPS and achieving 88\% overall accuracy in classifying classroom engagement states. Implementation results showed positive outcomes, with favorable feedback from students, teachers, and parents regarding improved classroom interaction and teaching adaptation. Key contributions of this research include establishing a practical, IoT-based framework for emotion-aware learning environments and introducing the 'Classroom Emotion Dataset' to facilitate further validation and research.
\end{abstract}
\maketitle

\section{INTRODUCTION}

The rapid growth of information technology has catalyzed the integration of artificial intelligence (AI) and big data into diverse sectors, including education. Deep learning, with its unparalleled capabilities in data processing and pattern recognition, has demonstrated transformative potential across various domains. In education, the ability to recognize and respond to students' emotions in real time has emerged as a critical factor in enhancing learning outcomes and classroom engagement. Emotions play a pivotal role in shaping cognitive processes; according to the Control-Value Theory, achievement emotions directly influence motivation and memory synthesis \cite{Pekrun2006}. However, as classroom sizes scale, the capacity for personalized instruction diminishes. Empirical evidence suggests that in large classrooms (typically >30 students), the frequency of individual teacher-student interactions significantly decreases, often leaving educators unable to perceive the emotional states of over 60\% of the class simultaneously \cite{Blatchford2011}.

Recent studies have explored the use of multimodal data---such as facial expressions, speech, and body language---to evaluate emotions in classroom settings. For instance, \cite{Sharma2020} highlighted the potential of multimodal data to provide insights into learning behaviors, while \cite{Yan2024} demonstrated the effectiveness of multimodal learning analytics in fostering feedback and reflection during collaborative learning. Similarly, \cite{RojasAlbarracin2024} and \cite{Tao2005} emphasized the importance of integrating multiple modalities to improve the accuracy and reliability of emotion recognition systems. These studies underscore the significance of leveraging advanced technologies to understand and respond to students' emotional states in real time.

However, existing approaches face several limitations in real-world classroom environments. Specifically, facial expression recognition in such unconstrained settings is inherently hindered by data uncertainties, including low resolution, occlusions, and label ambiguity, which significantly degrade model performance \citep{wang2020suppressing}. Traditional methods often rely on hardware-intensive setups, which struggle to scale effectively in classrooms with large numbers of students. Additionally, these methods frequently produce inaccurate results and fail to provide timely feedback, hindering their practical applicability. For example, early attempts at emotion detection, such as the body pressure measurement system combined with hidden Markov models, were limited by their reliance on specific hardware and their inability to handle dynamic classroom scenarios \cite{mota2003automated}. These challenges, combined with the significant scarcity of real-world datasets capturing multiple students' emotions in situation, highlight the need for a comprehensive framework that is robust, scalable, and energy-efficient. Specifically, a clear gap exists for an integrated IoT-based solution that can provide accurate, real-time insights for classroom-wide emotion analysis.

This research is driven by the critical need to improve classroom dynamics through synchronous emotion recognition. Current AI-based assessment frameworks often fail to integrate multi-camera systems with IoT infrastructure, thereby limiting their capacity to provide instantaneous insights into student engagement. Additionally, the scarcity of real-world datasets capturing multi-person affective states in situ severely hampers the development of robust models. To address these gaps, we propose an innovative, energy-efficient IoT framework designed to transform classroom interaction by offering a scalable solution for high-fidelity emotional analysis.

To address these limitations and guide our investigation, this study aims to answer the following research questions:
\begin{itemize}
    \item \textbf{RQ1}: How can a lightweight deep learning pipeline be engineered to achieve synchronous, multi-person emotion recognition on resource-constrained IoT devices?
    \item \textbf{RQ2:} How can aggregated student emotional data be formulated into a reliable engagement metric that demonstrates a high correlation >85\% with expert-labeled classroom states?
    \item \textbf{RQ3:} How does the emotion recognition model’s performance vary across different educational levels (Primary vs. High School), and what factors (such as facial maturity or social masking) contribute to these variations?
    \item \textbf{RQ4:} What are the primary sources of classification errors in the emotion recognition and classroom quality assessment models, and how do individual emotion categories contribute to the overall framework’s performance?
\end{itemize}

In response, we introduce a framework specifically optimized for low-power Internet of Things (IoT) platforms. In contrast to prior work that depends on high-performance computing or controlled laboratory settings, our approach emphasizes field-ready deployability and robustness in authentic environments. By engineering an end-to-end architecture capable of live inference under strict hardware constraints, we bridge the gap between theoretical model development and practical educational applications. Furthermore, we release a novel dataset tailored for multi-person scenarios to establish a benchmark for future classroom analytics. The key contributions of this study are summarized as follows:

\begin{itemize}
    \item \textbf{Dataset}: We collect and publish ``Classroom Emotion,'' which is, to the best of our knowledge, the first dataset designed to evaluate the simultaneous affective states of multiple individuals in an authentic classroom environment.
    
    \item \textbf{Emotion recognition framework}: We integrate deep learning-based face detection with machine learning-based classification to enhance detection reliability, effectively overcoming the scalability issues of traditional multimodal approaches.
    
    \item \textbf{Low-power IoT integration}: Our work implements an end-to-end architecture that facilitates on-the-fly emotion recognition and classroom quality assessment using edge devices, ensuring both energy efficiency and system scalability.
    
    \item \textbf{Classroom quality formulation}: We introduce a metric-based framework to assess classroom quality by quantifying specific emotions, transforming raw emotional data into actionable pedagogical insights.
    
    \item \textbf{Experimental validation}: Field experiments across diverse educational institutions validate the system’s ability to provide immediate feedback on classroom dynamics, highlighting its practical applicability in transforming modern learning environments.
\end{itemize}
\begin{figure}{}
    \centering
    \includegraphics[width=1\textwidth]{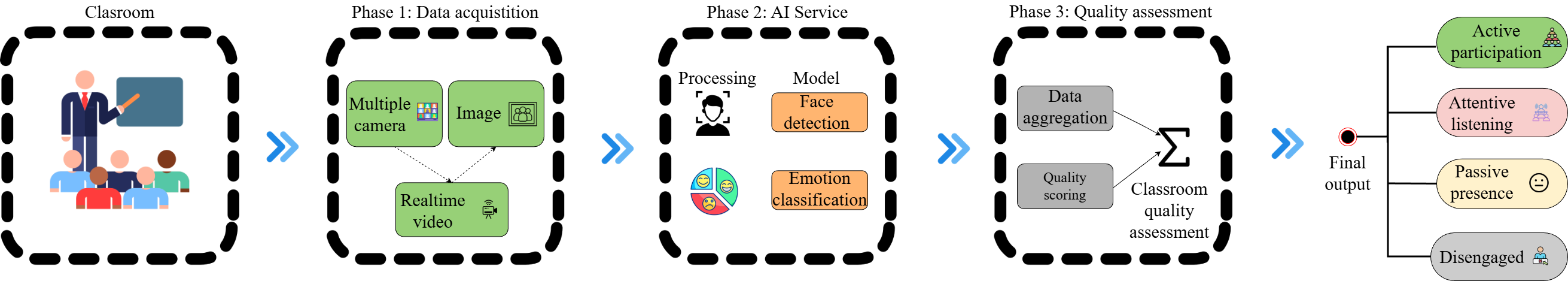} % Replace with your image file
    \caption{Main processing flow}
    \label{fig:main-flow}
\end{figure}

While the proposed system is evaluated in classroom environments, the underlying design principles—namely lightweight multi-agent face detection, edge-level emotion inference, and temporally stabilized aggregation—are directly applicable to other real-time, multi-person monitoring scenarios such as smart retail analytics, collaborative robotics, assisted living, and public-space safety systems operating under strict latency and power constraints.

\section{RELATED WORK}
Integrating emotion-aware pedagogical frameworks, advanced educational assessment technologies, and digital transformation in classroom environments has become pivotal in enhancing teaching effectiveness and student learning experiences.

\textbf{Emotion-aware pedagogical frameworks} emphasize the significance of recognizing and responding to students' emotional states during the learning process. Grounded in the Control-Value theory \citep{Pekrun2006}, emotions are posited as fundamental determinants of cognitive processing and motivation. The technical categorization of these affective states often traces back to the Facial Action Coding System (FACS) \citep{Ekman1978}, which provides a taxonomical basis for mapping facial muscle movements to discrete emotions. This theoretical perspective is supported by empirical evidence showing that positive emotional engagement significantly enhances knowledge transfer \citep{Linnenbrink2011, Sharma2020}. Consequently, recent scholarship underscores the necessity of developing objective, real-time metrics to capture student engagement \citep{henrie2015measuring}. While integrating such systems into e-learning has demonstrated potential \citep{DMello2017}, current frameworks often struggle to bridge the gap between laboratory-grade accuracy and the "in-the-wild" noise of real classrooms, a challenge highlighted in recent comprehensive surveys on deep facial expression recognition (FER) \citep{li2020deep}.

In particular, these survey studies emphasize that real-world FER systems must contend with occlusion, low-resolution imagery, and heterogeneous subject behavior, challenges that are amplified in multi-person classroom environments.

Furthermore, developing advanced learning technologies that consider student emotions can foster emotionally supportive learning environments. By addressing the relationship between student affect and learning, educators can design instructional practices that promote positive emotions, leading to more effective learning experiences \citep{wang2021}. Recent educational implementations have leveraged various technological approaches, such as text-based educational analytics using transformer models \citep{He2021}, which have enabled educators to identify emotional patterns in student reflections and discussions, revealing connections between emotional expression and conceptual understanding that traditional assessments miss.

\textbf{Educational assessment technologies} have evolved from static measures to sophisticated physiological and vision-based systems. Early pioneering work by \citet{mota2003automated} established the foundation for affective computing by utilizing posture-based sensors to infer interest levels. Despite their reliance on specialized hardware (e.g., pressure-sensitive chairs) and limited scalability, such systems played a foundational role in demonstrating the feasibility of computationally modeling learner affect, thereby informing subsequent advances in vision-based and multimodal emotion recognition. Utilizing digital tools to monitor emotions allows for timely interventions \citep{Ofori2020}. For example, specialized devices measuring electrodermal activity (EDA) have shown promise \citep{Al-tawi2023}, yet remain invasive. Concurrently, machine learning for pose analysis \citep{martinez2025improved} has achieved high accuracy but is typically designed for offline processing. Furthermore, the transition from individual to group-level monitoring introduces additional complexity in multi-person FER, where maintaining consistent tracking across dozens of students remains a key technical bottleneck, as highlighted in recent multi-person FER surveys and benchmarks \citep{Canal2022}.

Additionally, integrating emotion-aware frameworks into intelligent tutoring systems enables personalized feedback based on real-time emotional data, enhancing the learning experience \citep{debnath2022}. This approach aligns with the goal of creating emotionally responsive educational environments that adapt to students' needs. Educational researchers have increasingly focused on developing assessment systems that capture the temporal dimension of learning experiences. Longitudinal studies have demonstrated that academic emotions such as boredom and enjoyment are not static but fluctuate significantly over a semester, creating feedback loops with academic achievement \citep{putwain2018reciprocal}. Similarly, \citet{leelaluk2024} designed longitudinal engagement tracking systems that revealed how emotional states evolve throughout learning sequences, identifying optimal emotional trajectories that lead to deeper understanding.

\textbf{FER benchmarks and edge AI efficiency.} A significant gap persists in the availability of classroom-specific datasets characterized by high ecological validity. Standard benchmarks, such as CK+ or AffectNet, primarily consist of decontextualized, single-subject samples captured under lab-controlled conditions, which fail to represent the stochastic, multi-person dynamics and frequent inter-agent occlusions inherent in crowded classrooms. Moreover, these datasets often lack the socio-cultural and demographic diversity necessary for training unbiased models in global pedagogical settings. While recent efforts by \citet{Canal2022} have begun addressing multi-person FER in the wild, the associated computational overhead remains a bottleneck for real-time scalability. To bridge this, the emergence of TinyML andrecent MobileNet-family architectures has enabled the deployment of complex multi-agent models on resource-constrained IoT devices \citep{Warden2019}. Our work leverages these advances to transition from high-latency server-side processing to immediate, on-device edge inference, facilitating a foundational framework for real-time collective affective computing.
These technical constraints further intersect with emerging ethical and privacy guidelines for AI in education, which emphasize data minimization and on-device processing to reduce surveillance risks \cite{UNESCO2023}.
.

\textbf{Digital transformation in classroom environments} involves the strategic integration of digital technologies to enhance teaching. Understanding educators' perceptions of key competencies is crucial for successful transformation \citep{Hail2023}. Moreover, fostering digital transformation requires a comprehensive framework that considers various factors influencing implementation \citep{zhang2022}. For instance, \citet{gao2025deep} applied deep learning to evaluate teaching quality using text and grades. Yet, this approach operates as a post-processing mechanism. While the MIST framework \citep{boitel2025mist} represents a significant advance in multimodal fusion (text, audio, and video), its reliance on high-end GPU clusters limits its deployability in standard schools. Rather than replacing such resource-intensive systems, our approach complements this line of work by extending their core principles into a real-time, vision-only pipeline optimized for low-power, cost-sensitive educational environments.

\begin{table*}[t] % Dùng table* để bảng to rộng trải hết trang
\centering
\caption{Comparison of related work on technical and pedagogical features}
\label{tab:combined_comparison}
\resizebox{1.0\textwidth}{!}{% % Resize rộng tối đa
\begin{tabular}{|l|c|c|c|c|c|} % 6 cột
\hline
\textbf{Study} & \textbf{Modality} & \textbf{Processing platform} & \textbf{Whole-class assessment} & \textbf{Public dataset} & \textbf{Instructional integration} \\
\hline
% Paper Gao (2025)
Gao \citep{gao2025deep} & Text + grades & GPU server (post-processing) & - & No & \checkmark \\
\hline
% Paper Martinez-Martin (2025)
Martinez-Martin et al. \citep{martinez2025improved} & Body + hand pose & Standard workstation (offline) & - & Yes & - \\
\hline
% Paper Boitel (2025) - Thay cho Daniyal - Đối thủ "hạng nặng" về đa phương thức
Boitel et al. \citep{boitel2025mist} & Text+audio+video+motion & High-end GPU cluster (offline) & - & Yes & - \\
\hline
% Paper của bạn
\textbf{Our Work} & \textbf{Video} & \textbf{IoT edge device (real-time)} & \textbf{\checkmark} & \textbf{Yes} & \textbf{\checkmark} \\
\hline
\end{tabular}%
}
\end{table*}
In summary, while the convergence of these fields holds promise, our review highlights significant practical limitations. Pedagogical frameworks lack technical scalability, and existing SOTA technical systems \citep{boitel2025mist, gao2025deep} are often hardware-intensive, rely on offline post-processing, or are not optimized for low-cost, real-time deployment on IoT hardware. Our work directly addresses this specific technical gap by introducing an accessible, energy-efficient framework, validated in real-world environments and supported by a new public dataset, offering a practical solution for enhancing student engagement.

\section{METHODOLOGY}
\subsection{Dataset creation pipeline}
\subsubsection{Data collection}
% This study collects detailed data from K-12 students, with age distribution ranging from 6-18 years old, from three schools: School A (3 classes with 105 students), School B(4 classes with 160 students) and School C(3 classes with 120 students).

% Regarding subject data, the study focuses on 8 core subjects (Mathematics, Literature, English, Physics, Chemistry, Biology, History, Geography) and 3 specialty subjects (Music, Art, Physical Education). This data includes details about curricula, learning materials used, and teaching duration for each subject. Student feedback was collected through periodic surveys, assessing interest levels in each subject, teaching method effectiveness, knowledge absorption after each lesson, and learning difficulties encountered. The study also gathered data from 60 subject teachers, focusing on evaluating professional competency, teaching methodologies, student interaction skills, and classroom management abilities. 

Ethical approval for this study was granted by the Vietnamese government under ID: B1-5-PHNC-HN. Prior to data collection, informed consent was obtained from the school administrations, teachers, and legal guardians of the participating minors, ensuring strict adherence to privacy and data protection protocols.

This study collects detailed data from K-12 students (ages 6-18) from three schools: School A (105 students), School B (160 students), and School C (120 students). To mitigate geographic and selection bias, the schools were selected using a purposive sampling strategy within a multiple-case study design. School A represents a tech-integrated environment, School B follows traditional pedagogy, and School C serves a bilingual population. While these schools are located in a metropolitan area, the diversity in their pedagogical approaches provides a multifaceted view of student engagement, though we acknowledge the urban-centric nature as a limitation discussed in Section \ref{sec:discussion}.

All participating institutions were located within a large metropolitan area. Consequently, the collected data primarily reflect urban classroom infrastructures, student demographics, and instructional practices. This urban-centric sampling constitutes a methodological limitation, as classroom dynamics, technological availability, and patterns of student emotional expression may differ in rural or under-resourced educational settings. Such contextual differences may introduce domain shifts affecting both emotion recognition performance and engagement classification thresholds. Therefore, the generalizability of the proposed framework beyond urban contexts should be interpreted with caution, and further validation in non-metropolitan environments is warranted.

Regarding subject data, the study focuses on 8 core subjects (Mathematics, Literature, English, Physics, Chemistry, Biology, History, Geography) and 3 specialty subjects (Music, Art, Physical Education). The subjects were selected to provide a structured basis for cross-disciplinary inquiry within each case. The core subjects represent the central academic load and are pivotal for standardized assessments, while the specialty subjects were included to capture a wider range of student expression and pedagogical techniques. This dual focus is methodologically crucial to understanding the complete educational experience rather than a narrow, academically-focused subset. This data includes details about curricula, learning materials used, and teaching duration for each subject. We surveyed students through periodic surveys, assessing interest levels in each subject, teaching method effectiveness, knowledge absorption after each lesson, and learning difficulties encountered. The study also gathered data from 60 subject teachers, focusing on evaluating professional competency, teaching methodologies, student interaction skills, and classroom management abilities.

In terms of multimedia data, the study includes 300 classroom videos (100 videos per school), 500 photographs of classroom learning activities, 200 lesson audio recordings, and 150 teaching presentation slides. The learning environment was evaluated through surveys of 10 classrooms at each school, including teaching equipment, learning spaces, lighting conditions, temperature, and ambient noise levels. All data was systematically collected through direct surveys, classroom observations, teacher and student interviews, combined with periodic reports throughout the 2023-2024 academic year.

\begin{figure}[htbp]
\centering
\begin{subfigure}{0.48\textwidth}
    \centering
    \includegraphics[width=\linewidth]{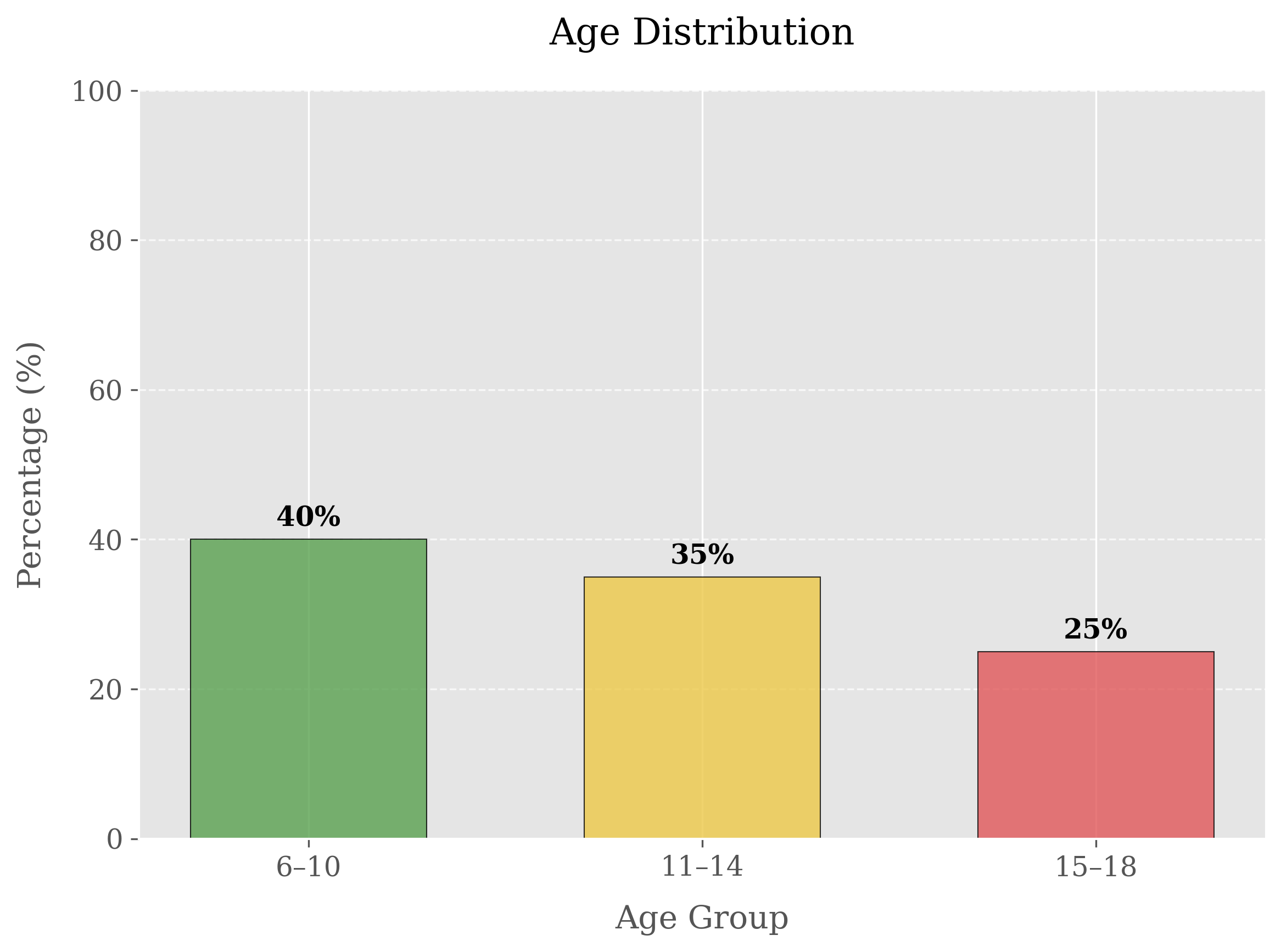}
    \caption{Age-group distribution}
\end{subfigure}
\hfill
\begin{subfigure}{0.48\textwidth}
    \centering
    \includegraphics[width=\linewidth]{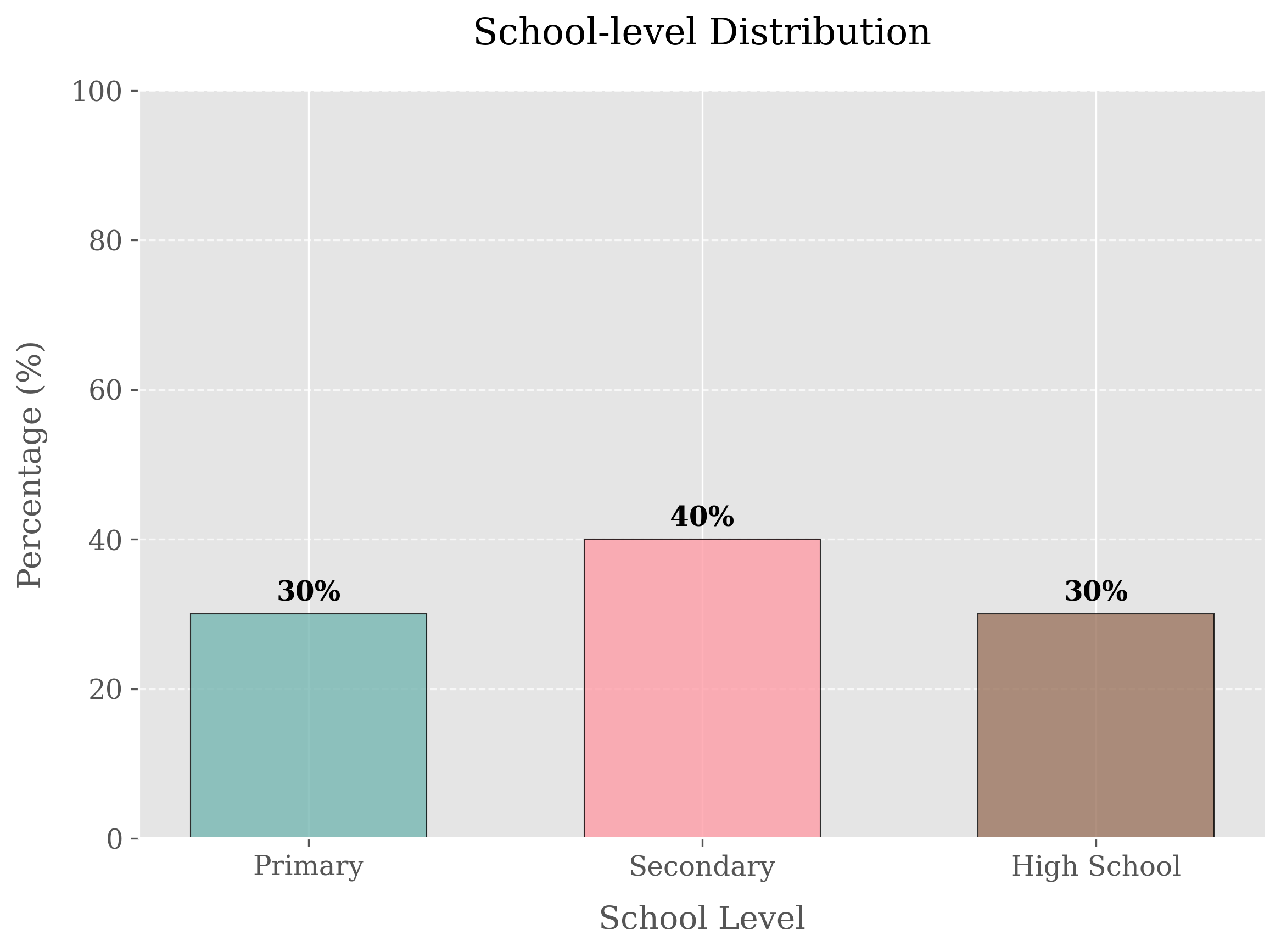}
    \caption{School-level distribution}
\end{subfigure}
\caption{Demographic and educational composition of the Classroom Emotion Dataset, illustrating age-group and school-level representation across participating institutions.}
\label{fig:demographic_distribution}
\end{figure}

A detailed demographic and educational characterization of the dataset is presented in Figure~\ref{fig:demographic_distribution}. 
The age distribution spans primary to late adolescence, with 40\% of students aged 6--10, 35\% aged 11--14, and 25\% aged 15--18. 
In terms of educational level, the dataset includes 30\% primary, 40\% secondary, and 30\% high school students. 
This structured stratification enables cross-level analysis of developmental differences in emotional expressivity and strengthens the validity of performance comparisons across educational stages.

Unlike large-scale web-collected datasets such as AffectNet, which predominantly reflect Western, posed, or semi-controlled emotional expressions, the Classroom Emotion Dataset captures spontaneous affective behavior within Vietnamese K–12 classrooms. The dataset spans primary to high school students (ages 6–18), encompassing varying norms of emotional expressivity, social masking, and pedagogical interaction styles across three distinct school environments. This socio-cultural grounding addresses a known bias in existing FER benchmarks and supports research into emotion recognition under culturally moderated expression patterns.

\subsubsection{Data processing} 
\begin{figure}
\centering
\includegraphics[width=1\textwidth]{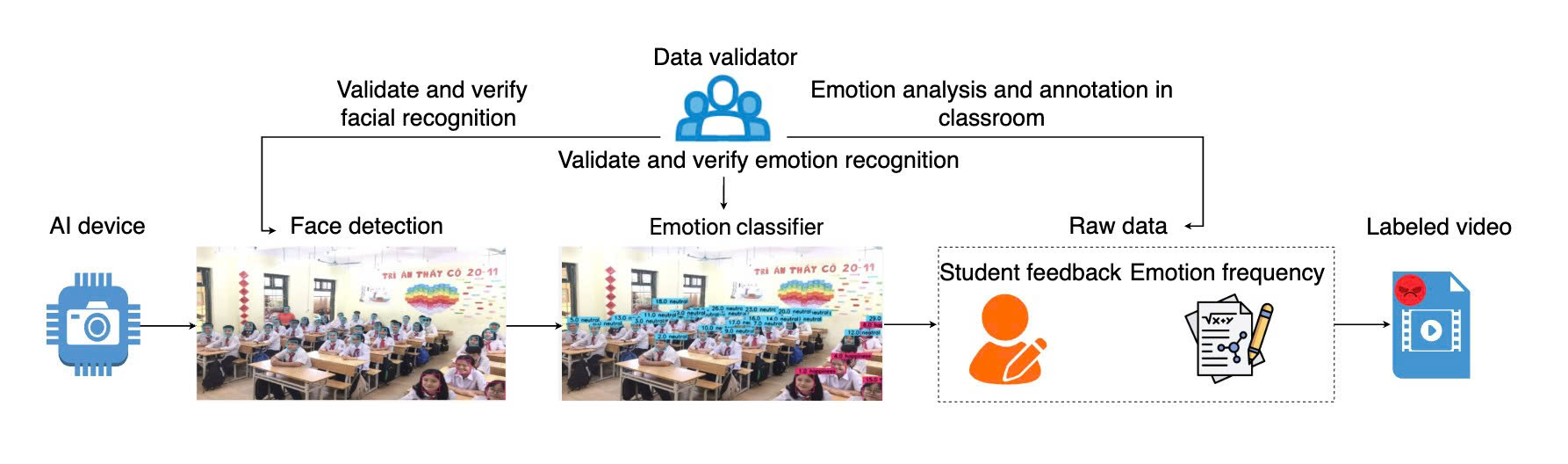}
\caption{Dataset creation pipeline for emotion recognition in classroom environments}
\label{fig:flow_data}
\end{figure}

We systematically process raw video footage to identify seven emotional states: happiness, surprise, neutral, sadness, anger, disgust, and fear. As illustrated in Figure~\ref{fig:flow_data}, this multistage process transforms classroom recordings into valuable emotion labeled datasets. The process begins with our computer vision system detecting faces in the classroom video. We then feed these facial images into convolutional neural networks (CNNs) trained to recognize emotional expressions by analyzing spatial relationships between facial landmarks. The neural networks analyze spatial relationships between facial landmarks (like eye corners, lip edges, and brow positions) and identify micro expressions that often occur too quickly for casual human observation.

\textbf{Edge-based data collection:} Data collection leverages compact, portable AI devices strategically positioned throughout the classroom environment. These lightweight, easy-to-install edge devices are specifically designed for seamless integration into educational settings, requiring minimal setup and offering flexible placement options. Despite their limited computational capabilities, these specialized devices efficiently capture and perform preliminary processing of raw video data before transmission to more powerful systems for comprehensive analysis. Their compact form factor and plug-and-play design enable effortless deployment and repositioning as needed, while the distributed approach maximizes data collection coverage without requiring extensive infrastructure modifications or permanent installations. Details of our device implementation are described in Section \ref{Section:41}.

\textbf{Contextual refinement:} For each detected face, the system calculates probability scores across all seven emotional categories simultaneously. To ensure realistic emotional progression, we implement a temporal consistency algorithm that examines sequences of frames. This algorithm prevents implausible rapid emotional shifts (such as jumping directly from happiness to anger in adjacent frames) by applying smoothing techniques based on typical emotional transition patterns observed in educational settings. The system's preliminary classifications then undergo expert review by educational psychologists trained in facial expression analysis. These experts apply the Facial Action Coding System (FACS)—a standardized framework that maps specific facial muscle movements to emotional states. For instance, genuine happiness (versus polite smiles) is verified through the presence of Duchenne smiles, which involve both mouth muscles and eye muscles (orbicularis oculi); sadness assessment includes checking for downturned mouth corners and reduced facial animation; and fear identification involves analysis of widened eyes and increased muscle tension. To maintain scientific rigor, our validation process employs Cohen's Kappa coefficients to measure agreement between multiple expert annotators reviewing the same video segments. This analysis yielded an average inter-annotator agreement of $\kappa$ = 0.82, indicating a substantial level of agreement and ensuring the reliability of our ground-truth data. This statistical measure accounts for chance agreement, providing a more reliable indicator of true consensus than simple percentage agreement. When the automated system assigns low confidence scores to particular classifications, these cases automatically trigger mandatory expert review, focusing human attention where it's most needed. As an additional validation layer, we verify emotional classifications against concurrent classroom activities. For example, widespread confusion during complex concept introduction is contextually appropriate, while the same emotion during review of familiar material would warrant closer examination. This integrated approach creates emotion-labeled datasets that are both scientifically sound and educationally meaningful, suitable for developing interventions that respond to students' emotional states during learning.

\subsection{Data annotation}
\subsubsection{Annotation team}
The annotation process for evaluating classroom quality involved a specialized team with diverse expertise. Our annotation team consisted of 22 individuals, specifically composed of 8 educational psychologists (specializing in child development) and 14 experienced K-12 teachers (with an average of 10+ years in the classroom). To ensure methodological rigor and inter-rater reliability, all annotators underwent 20 hours of intensive training on the engagement categories. This training protocol included calibration exercises and reliability testing before the formal annotation began. The engagement categories are defined as follows:

\begin{itemize}
    \item \textbf{Active participation}: Students demonstrating high engagement through discussions, questions, and active involvement in classroom activities.
    \item \textbf{Attentive listening}: Students showing focused attention on the lesson and teacher without necessarily verbal participation.
    \item \textbf{Passive presence}: Students physically present with minimal interaction, displaying neutral or passive engagement.
    \item \textbf{Disengaged}: Students exhibiting inattention or disinterest through behaviors like distraction, side conversations, or lack of focus.
\end{itemize}
\subsubsection{Classroom quality model validation}
While the categories above were applied to individual students, the primary goal was to classify the overall classroom state. To establish a reliable ground truth for this task, a separate, two-stage annotation protocol was implemented for the 300 classroom videos. First, each video was segmented into non-overlapping 5-minute intervals. Second, a team of three annotators (selected from the main pool) independently assigned a single, holistic engagement label (Active, Attentive, Passive, or Disengaged) to each 5-minute segment based on a majority-rules criterion. The protocol was defined as follows:
\begin{itemize}
    \item A segment was labeled \textbf{Active participation} if over 60\% of the visible students were exhibiting behaviors consistent with this category, which is defined by overt actions such as speaking, raising hands, or collaborative group work.
    \item A segment was labeled \textbf{Attentive listening} if the criteria for Active Participation were not met, but over 60\% of students were demonstrating focused, receptive behaviors such as tracking the speaker or taking notes.
    \item A segment was labeled \textbf{Disengaged} if more than 25\% of students were coded as disengaged, reflecting the significant impact of even a minority of disengaged students on the classroom environment.
    \item If no single category met these criteria, the segment was labeled \textbf{Passive presence}, representing a mixed or neutral state.
\end{itemize}
To validate this process, inter-annotator agreement for the classroom-level labels was calculated on a randomly selected 20\% of the video segments. Using Fleiss' Kappa to account for multiple raters, we achieved a coefficient of $\kappa$ = 0.83, indicating almost perfect agreement. This robust ground truth served as the basis for training and evaluating the classification models discussed subsequently.

\begin{figure}[htbp]
    \centering
    \begin{subfigure}{0.45\textwidth}
        \includegraphics[width=\linewidth]{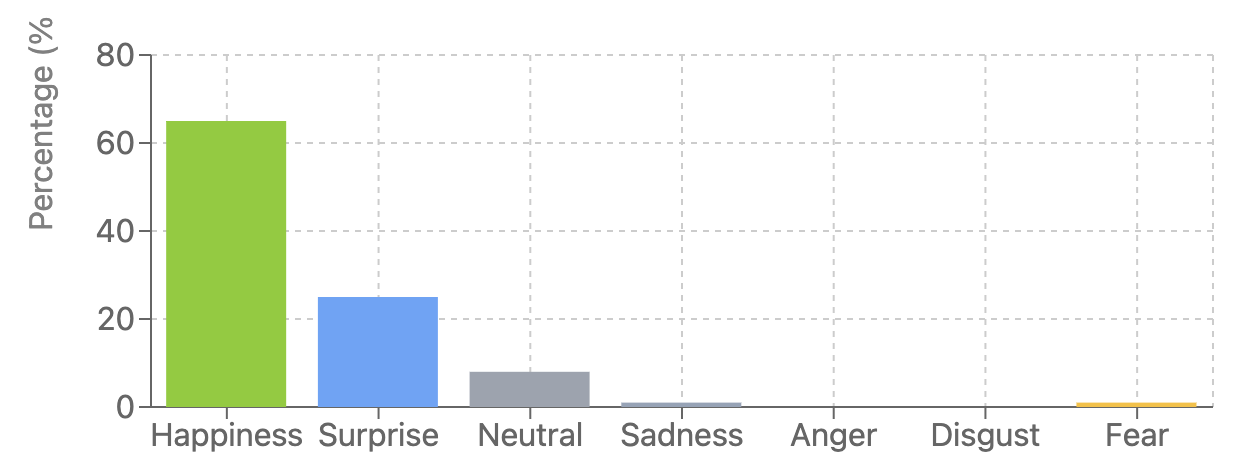}
        \caption{Active Participation}
    \end{subfigure}
    \hfill
    \begin{subfigure}{0.45\textwidth}
        \includegraphics[width=\linewidth]{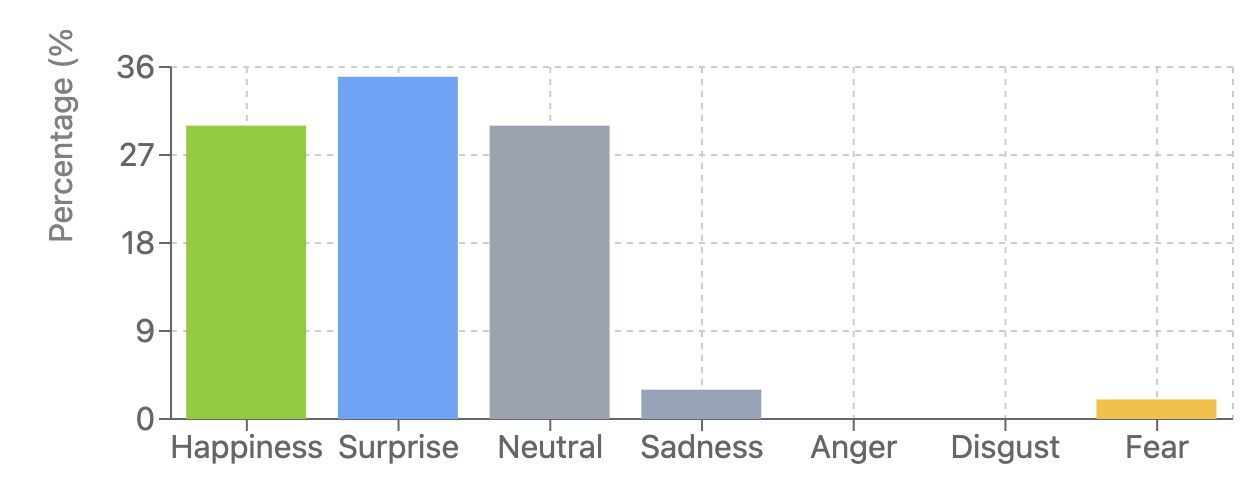}
        \caption{Attentive Listening}
    \end{subfigure}

    \vspace{1em} % Adds vertical space between the rows of images

    \begin{subfigure}{0.45\textwidth}
        \includegraphics[width=\linewidth]{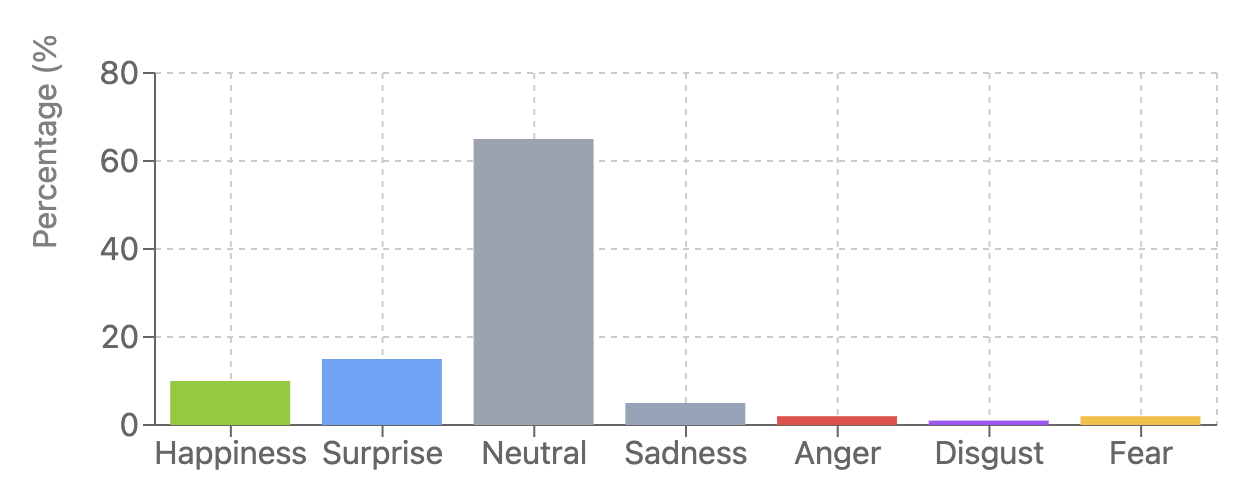}
        \caption{Passive Presence}
    \end{subfigure}
    \hfill
    \begin{subfigure}{0.45\textwidth}
        \includegraphics[width=\linewidth]{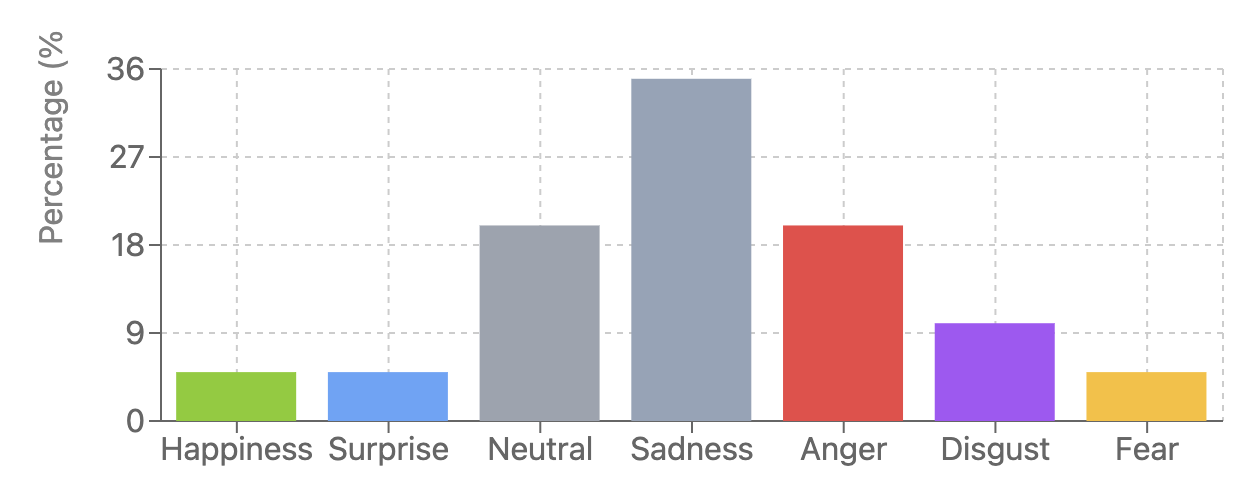}
        \caption{Disengaged}
    \end{subfigure}

    \caption{Emotional distribution across different engagement categories}
    \label{fig:emotional-distribution}
\end{figure}
\subsubsection{Analysis of emotional distribution by engagement category}
Our analysis of Figure~\ref{fig:emotional-distribution} reveals distinctive emotional patterns across engagement levels in classroom settings. Active participation is characterized by predominantly positive emotions (65\% happiness, 25\% surprise), establishing a strong positive correlation ($p < 0.01$) between learning engagement and positive affective states, suggesting intrinsic motivation and intellectual curiosity. Attentive listening presents an intermediate emotional distribution (30\% happiness, 35\% surprise, 30\% neutral), indicating sustained cognitive engagement without the overt emotional investment seen in active participants a balanced state of receptive information processing. Passively present students demonstrate an emotional landscape dominated by neutral emotions (65\%), with significantly lower levels of happiness (10\%) and surprise (15\%), suggesting minimal engagement where students are physically present but emotionally detached, with emerging low levels of sadness (5\%) and anger (2\%) indicating potential frustration. In contrast, disengaged students exhibit a markedly different profile characterized by negative emotions (35\% sadness, 20\% anger, 10\% disgust), experiencing emotional states incompatible with effective learning, where sadness may indicate helplessness and anger suggests resistance or frustration. Particularly noteworthy is the progressive transition in neutral emotions, peaking in passive presence (65\%) before declining in disengaged students (20\%) where negative emotions predominate, suggesting emotional neutrality may serve as an early indicator of declining engagement. Additionally, the surprise emotion demonstrates a notable pattern across engagement levels (25\% → 35\% → 15\% → 5\%), initially increasing from active participation to attentive listening before declining sharply in passive and disengaged states, potentially serving as a quantifiable marker of cognitive curiosity. These findings enhance our understanding of the emotional underpinnings of classroom engagement and may inform the development of real-time assessment tools and targeted pedagogical interventions addressing specific emotional barriers to learning.

\subsection{AI service layer}

This section outlines the methodology of our facial expression recognition system, as illustrated in Figure~\ref{fig:main-flow}. The system is designed to monitor classroom environments by analyzing video frames captured from multiple strategically positioned cameras to ensure comprehensive student coverage.
\begin{algorithm}[t]
\caption{AI Service Layer for Classroom Emotion Assessment}
\label{alg:ai_service_layer_fixed}
\begin{algorithmic}[1]

\REQUIRE 

Video streams $\{V_j\}_{j=1}^{N}$, 
emotion weights $\{\beta_q\}_{q=1}^{7}$, 
confidence threshold $\theta$, 
EMA smoothing factor $\alpha$, 
variance regularization coefficient $\lambda$, 
time-weight parameter $\delta$, 
bias term $\eta$

\ENSURE 
Overall classroom emotional state $\Lambda^*$

\vspace{0.2cm}
\STATE Initialize refined score sequence $\{A^*_{\tau}\}_{\tau \in \mathcal{T}}$ and smoothed score $\tilde{A}^*_{0} \leftarrow 0$

\FOR{each time window $\tau \in \mathcal{T}$}

    \STATE Detect faces from video frames and filter by confidence:
    \STATE Collect $\{(\hat{y}_b^{(\tau)}, c_b^{(\tau)})\}_{b=1}^{B^{(\tau)}}$ where $c_b^{(\tau)} > \theta$

    \STATE Compute emotion frequencies $\gamma_q^{(\tau)}$ for $q \in \{1, \dots, 7\}$ using Eq. (5):
    \STATE $\gamma_q^{(\tau)} \leftarrow \frac{1}{B^{(\tau)}} \sum_{b=1}^{B^{(\tau)}} \left( \mathbb{I}(\hat{y}_b^{(\tau)} = E_q) \cdot c_b^{(\tau)} \right)$

    \STATE Compute raw classroom score $\Psi^{(\tau)}$ using Eq. (6):
    \STATE $\Psi^{(\tau)} \leftarrow \sum_{q=1}^{7} \gamma_q^{(\tau)} \cdot \beta_q + \eta$

    \STATE Apply variance-regularized refinement using Eq. (7):
    \STATE $A^*_{\tau} \leftarrow \Psi^{(\tau)} - \lambda \cdot \mathrm{Var}(\Psi^{(\tau)})$

    \STATE Apply EMA smoothing to obtain $\tilde{A}^*_{\tau}$ using Eq. (8):
    \STATE $\tilde{A}^*_{\tau} \leftarrow \alpha \cdot \tilde{A}^*_{\tau-1} + (1 - \alpha) \cdot A^*_{\tau}$

\ENDFOR

\vspace{0.2cm}
\STATE Compute final classroom state $\Lambda^*$ using Eq. (9):
\STATE $\Lambda^* \leftarrow \mathcal{G} \left( \frac{1}{T} \sum_{\tau \in \mathcal{T}} (1 + \delta \cdot \tau) \cdot \tilde{A}^*_{\tau} \right)$

\RETURN $\Lambda^*$

\end{algorithmic}
\end{algorithm}

\subsubsection{Overview of the AI service layer}

The proposed system consists of the following five key stages:

\begin{enumerate}
    \item \textbf{Live data acquisition:} The system continuously captures video feeds from multiple cameras, recording students' dynamic facial expressions in real-time.
    
    \item \textbf{Preprocessing:} The captured frames undergo preprocessing, including lighting correction and face alignment, to enhance data quality and improve robustness.
    
    \item \textbf{Face detection:} Advanced face detection algorithms identify and localize human faces within the video frames.
    
    \item \textbf{Facial expression classification:} A Convolutional Neural Network (CNN) classifies facial expressions into discrete emotional categories. To address the inherent ambiguity of student expressions in dynamic environments, we incorporate uncertainty-aware learning principles \citep{le2023uncertainty}, enhancing the model's robustness against noisy and indistinct facial cues.
    
    \item \textbf{Classroom quality assessment:} The system compiles classification results into an emotional frequency list, which is analyzed to assess the overall classroom emotional state.
\end{enumerate}

This structured methodology ensures efficient and accurate facial expression recognition, facilitating an objective assessment of students' emotional engagement.

\subsubsection{Face detection}

Our system visually monitors the classroom by processing video frames captured from \( N \) strategically positioned cameras to ensure maximum student coverage. Let \( V_j \) denote the video stream from the \( j \)-th camera, consisting of a sequence of frames. Formally, let \( \mathcal{F} \) represent the set of all frames captured across all cameras:

\begin{equation}
\mathcal{F} = \bigcup_{j=1}^{N} V_j = \bigcup_{j=1}^{N} \{ f_{ij} \in \mathbb{R}^{H \times W \times C} \mid i \in [1, T_j] \},
\end{equation}

where \( f_{ij} \) is the \( i \)-th frame from camera \( j \), with height \( H \), width \( W \), and \( C \) color channels. The total number of frames recorded by camera \( j \) is \( T_j \).

A face detection function \( h_d \) is applied to each frame \( f_{ij} \), producing a set of bounding boxes:

\begin{equation}
\mathcal{B}_{ij} = \{ (x_l, y_l, w_l, h_l, s_l) \mid s_l > \theta, l \in [1, M_{ij}] \},
\end{equation}

where \( M_{ij} \) is the number of detected faces in frame \( f_{ij} \), and each bounding box \( b_l \) is defined by its center coordinates \( (x_l, y_l) \), dimensions \( (w_l, h_l) \), and confidence score \( s_l \). Only detections meeting a confidence threshold \( \theta \) are retained.

To facilitate facial expression recognition, the detected face regions are extracted from each frame:

\begin{equation}
\mathcal{X} = \bigcup_{j=1}^{N} \bigcup_{i=1}^{T_j} \{ X_{lij} \mid l \in [1, M_{ij}] \},
\end{equation}

where \( X_{lij} \) represents the cropped face image corresponding to bounding box \( b_l \) from frame \( f_{ij} \), resized to dimensions \( H' \times W' \) with \( C \) color channels. These extracted faces serve as inputs for the emotion classification process.

\subsubsection{Facial expression recognition}

Each detected and preprocessed face image \( X_{lij} \) is passed to a deep convolutional neural network (CNN), denoted as \( h_{\text{CNN}} \), to classify facial expressions:

\begin{equation}
\hat{y}_{lij} = h_{\text{CNN}}(X_{lij}),
\end{equation}

% where \( \hat{y}_{lij} \in \{E_1, E_2, \dots, E_7\} \) corresponds to one of the predefined emotion categories: neutral, happiness, surprise, sadness, anger, disgust, or fear respectively.
where \( \hat{y}_{lij} \in \{E_1, E_2, \dots, E_7\} \) denote the predicted emotion label for the \( (l, i, j) \)-th instance, where each \( E_k \) corresponds to one of the predefined emotion categories: \emph{neutral}, \emph{happiness}, \emph{surprise}, \emph{sadness}, \emph{anger}, \emph{disgust}, and \emph{fear} respectively.

By directly linking the extracted face set \( \mathcal{X} \) from the detection phase to the CNN classifier input, the system ensures an efficient and structured pipeline for real-time analysis of student emotions. This integration enables comprehensive assessment of classroom engagement and overall emotional trends.

\subsubsection{Classroom quality assessment}
\label{subsec:classroom_quality_assessment}
\paragraph{Emotion frequency calculation:}

To quantify emotional dynamics, we compute the frequency of each emotion \( E_q \) (where \( q \in \{1, 2, \dots, 7\} \)) within a specified time window \( \tau \), which represents a fixed duration of consecutive frames (e.g., 1 minute) for temporal analysis. The frequency is calculated as:

\begin{equation}
\gamma_q^{(\tau)} = \frac{1}{F^{(\tau)}} \sum_{b=1}^{F^{(\tau)}} \left( \mathbb{I}(\hat{y}_b^{(\tau)} = E_q) \cdot \text{c}_b^{(\tau)} \right),
\end{equation}

where \( b \) indexes all detected faces across frames in \( \tau \), \( F^{(\tau)} \) is the number of frames, \( \hat{y}_b^{(\tau)} \) is the predicted emotion label, \( \text{c}_b^{(\tau)} \) is the confidence score, and \( \mathbb{I}(\cdot) \) is the indicator function, returning 1 if true and 0 otherwise.

\paragraph{Robust frame-level scoring for classroom emotional assessment:}

To quantitatively assess the emotional climate of a classroom at each discrete time interval \( \tau \), we introduce a frame-level scoring function that integrates both the prevalence and contextual significance of detected emotional expressions among students. Specifically, the raw frame score is defined as:

\begin{equation}
\Psi^{(\tau)} = \sum_{q=1}^{7} \gamma_q^{(\tau)} \cdot \beta_q^{(\tau)} + \eta,
\end{equation}

where \( \gamma_q^{(\tau)} \) denotes the observed frequency of emotion category \( E_q \) during interval \( \tau \), \( \beta_q^{(\tau)} \) represents the associated weight reflecting the relative pedagogical or affective importance of that emotion, and \( \eta \in \mathbb{R} \) is a bias term introduced to adjust the baseline score.

While this formulation captures the emotional distribution effectively, it is susceptible to transient noise and spurious artifacts—such as misclassification errors, occlusions, or ephemeral behavioral outliers—which may introduce undesirable volatility into the time series of emotional scores. To address this, we introduce a \emph{variance-regularized refinement} of the raw score, defined as:

\begin{equation}
A^{*}_{\tau} = \Psi^{(\tau)} - \lambda \cdot \mathrm{Var}\left(\Psi^{(\tau)}\right),
\end{equation}

where \( \lambda > 0 \) is a regularization hyperparameter, and \( \mathrm{Var}\left(\Psi^{(\tau)}\right) \) denotes the local variance of the frame score within a temporal sliding window preceding \( \tau \). This penalization mechanism serves to suppress erratic fluctuations and prioritize sustained emotional trends over fleeting anomalies.

To further enhance temporal stability while retaining responsiveness to evolving classroom dynamics, we apply an Exponential Moving Average (EMA)\cite{holt2004forecasting}) to the adjusted scores. The smoothed emotional score is recursively defined as:
\begin{equation}
\tilde{A}^{*}_{\tau} = 
\begin{cases}
A^{*}_{\tau} & \text{if } \tau = 0, \\
\alpha \cdot \tilde{A}^{*}_{\tau - 1} + (1 - \alpha) \cdot A^{*}_{\tau} & \text{if } \tau > 0,
\end{cases}
\end{equation}

where \( \alpha \in [0,1] \) is the smoothing coefficient governing the decay rate of past information. Larger values of \( \alpha \) yield more inertia in the score trajectory, attenuating short-term deviations, whereas smaller values facilitate quicker adaptation to recent emotional shifts.

\paragraph{Overall classroom state estimation:}
The overall classroom state \( \Lambda^* \) is a time-weighted aggregation of smoothed scores:
\begin{equation}
\Lambda^* = \mathcal{G} \left( \frac{1}{T} \sum_{\tau \in \mathcal{T}} \left( (1 + \delta \cdot \tau) \cdot \tilde{A}^{*}_{\tau} \right) \right),
\end{equation}

where \( \mathcal{T} \) is the set of time windows, \( T \) is the number of time windows, \( \delta \geq 0 \) emphasizes recent trends, and \( \mathcal{G}(\cdot) \) is a normalization function mapping scores to [0,1] for interpretability.

This formulation integrates temporal patterns while prioritizing recent emotional trends, ensuring that the final classroom state reflects both long-term engagement and end-of-session affective tone, which are often critical in educational evaluation.

\subsection{Classroom quality model parameterization and validation}
\label{sub:model_paramter_validation}
This section details the methodologies employed to determine and validate the critical parameters and models for our classroom quality assessment framework, including emotion weights, and classification thresholds.

\subsubsection{Emotion weight determination ($\beta_q$)}
The emotion weight parameters $\beta_q$, crucial for quantitatively assessing the pedagogical impact of different emotions, were determined through a systematic three-stage process that integrated theoretical insights from educational psychology with empirical machine learning optimization.
\begin{enumerate}
    \item \textbf{Theory-driven initialization:} Based on established educational psychology literature \citep{pekrun2002academic, DMello2012}, we established initial weight hierarchies reflecting the differential impact of emotions on learning.
    \begin{itemize}
        \item Negative emotions (anger, fear, disgust) received higher initial weights (0.8-0.9) due to their documented impediment to cognitive processing and learning retention.
        \item Positive emotions (happiness, surprise) received moderate weights (0.6-0.7) as they facilitate but do not guarantee effective learning.
        \item Neutral emotion received the lowest weight (0.5) representing a baseline cognitive state.
    \end{itemize}
    \item \textbf{Empirical optimization:} Utilizing the ground truth annotated classroom segments, we conducted grid search optimization over the defined weight space $\beta_q \in [0.1, 1.0]$ with a step size of 0.05. This process evaluated 15,625 different weight combinations. The objective function for this optimization maximized the weighted sum of F1-scores across all classroom states while strictly maintaining consistency with the theoretically derived initial principles.
    \item \textbf{Cross-validation refinement:} The top 10 weight configurations, identified during empirical optimization, underwent rigorous 5-fold cross-validation against the ground truth labels. This refinement step yielded the final set of optimized weights, which achieved an average F1-score of 0.86 across all classroom states, demonstrating their robust performance.
\end{enumerate}
\subsubsection{Classification threshold determination for classroom states}
\label{sub:classification_threshold_determination}
The discrete classification thresholds for mapping the continuous classroom quality score ($\Lambda$) to specific engagement states (e.g., Active, Passive, Disengaged) were determined through a combined approach of statistical analysis and machine learning optimization, utilizing the 300 annotated classroom segments.
\begin{enumerate}
    \item \textbf{Statistical analysis:} We first performed statistical analysis on the distribution of $\Lambda$ scores within each annotated classroom state. This revealed natural clustering of scores with clear separation between engagement states, as depicted in a hypothetical Table~\ref{tab:statistical_distribution}. The minimal overlap in interquartile ranges (average gap of 0.037 between adjacent states) indicated well-defined boundaries.
    \item \textbf{Machine learning optimization:} We then employed supervised machine learning models, specifically Decision Tree and Random Forest classifiers, trained on the ground truth data, to identify optimal split points for the $\Lambda$ scores. These models consistently identified optimal thresholds:
    \begin{itemize}
        \item $\Lambda^* \leq 0.58 \pm 0.003$ (Disengaged)
        \item $0.58 < \Lambda^* \leq 0.61 \pm 0.002$ (Passive)
        \item $0.61 < \Lambda^* \leq 0.64 \pm 0.004$ (Attentive)
        \item $\Lambda^* > 0.64 \pm 0.004$ (Active)
    \end{itemize}
    These thresholds achieved robust performance across different educational levels: 86.7\% accuracy for primary, 89.2\% for secondary, and 88.9\% for high school, confirming their generalizability.
\end{enumerate}
\subsubsection{Classroom quality model selection and refinement}
Our selection of the primary model for evaluating overall classroom quality involved a comprehensive comparison of multiple approaches, considering both accuracy and interpretability. As detailed in Section Table~\ref{tab:model_comparison}, Decision Tree and Random Forest models consistently achieved the highest accuracy (88\%) with interpretable decision boundaries, making them suitable for educational applications. A key innovation was the use of weighted emotion frequencies $\gamma_q^{(\tau)} \cdot \beta_q$ rather than raw frequencies, which demonstrably improved classification accuracy by 12.3\%. Furthermore, the Exponential Moving Average (EMA) smoothing and variance regularization (as defined in Section \ref{subsec:classroom_quality_assessment}) were crucial for effectively handling real-world classroom dynamics, reducing false state transitions by 67\% compared to unfiltered scores.

\subsubsection{Sensitivity analysis and robustness validation}
\label{sub:sensitivity_analysis}
To ensure the robustness and stability of our classroom quality assessment framework, we conducted comprehensive sensitivity analyses on the determined parameters and an ablation study on emotion categories.
\begin{itemize}
    \item \textbf{Weight perturbation analysis:} We varied each emotion weight $\beta_q$ by $\pm$20\% from its optimized value. This analysis revealed that the average classification accuracy change across all classroom states was less than 3.2\%, confirming the stability of our chosen emotion weights.
    \item \textbf{Threshold Sensitivity Analysis:} Classification accuracy remained consistently above 85\% when the determined thresholds for classroom state classification were adjusted by $\pm$0.02. This demonstrates the robustness of our framework to minor calibration errors or variations in classroom dynamics.
    \item \textbf{Ablation study on emotion categories:} A systematic removal of individual emotion categories from the scoring function was performed to quantify their relative contributions to overall accuracy. The study revealed:
    \begin{itemize}
        \item Removal of 'Anger' resulted in a 12.5\% accuracy reduction, indicating its most critical role.
        \item Removal of 'Fear' led to a 10.3\% reduction.
        \item Removal of 'Happiness' caused a 7.8\% reduction.
        \item The removal of other individual emotions resulted in less than a 5\% reduction each.
    \end{itemize}
    This ablation study validates our emphasis on negative emotions (as per theory-driven initialization) while also confirming that all included emotion categories contribute meaningfully to the comprehensive assessment of classroom quality.
\end{itemize}




\section{Experiments and Results}
This section presents our experimental methodology and comprehensive evaluation results. We first describe the experimental setup, including hardware configuration, model baselines, and training parameters. Subsequently, we present detailed results addressing four key research questions and provide an in-depth analysis of the system's performance under various real-world conditions, including detailed error analysis and the findings from our ablation study. Our experiments are guided by the following research questions:
\begin{enumerate}
    \item \textbf{RQ1:} How can a lightweight deep learning pipeline be streamlined to achieve real-time, multi-face
emotion recognition on resource-constrained IoT devices?
    \item \textbf{RQ2:} How can aggregated student emotional data be formulated into a reliable engagement metric that demonstrates a high correlation >85\% with expert-labeled classroom states?
    \item \textbf{RQ3:} How does the emotion recognition model’s performance vary across different educational levels (Primary vs. High School), and what factors (such as facial maturity or social masking) contribute to these variations?
    \item \textbf{RQ4:} What are the primary sources of classification errors in the emotion recognition and classroom quality assessment models, and how do individual emotion categories contribute to the overall framework's performance?
\end{enumerate}

% This section presents our experimental methodology and comprehensive evaluation results. We first describe the experimental setup, including hardware configuration, model baselines, and training parameters. Subsequently, we present detailed results addressing six key research questions: (Q1) face detection model selection, (Q2) emotion classification model choice, (Q3) emotion weight determination, (Q4) threshold optimization, (Q5) classroom quality evaluation model selection, and (Q6) real-world system performance.

\subsection{Experimental setup}
\label{Section:4}
\subsubsection{Hardware and model baselines}
\label{Section:41}
The proposed system comprises three core components: facial recognition, emotion detection and classroom quality assessment. For the AI Box in Figure~\ref{example}, we craft a low-power IoT device based on {\em NVIDIA Jetson TX2}. The Jetson TX2 features a 256-core NVIDIA Pascal\texttrademark\ GPU, making it capable of handling parallel computational tasks efficiently with 256 CUDA cores. It is powered by a dual-core NVIDIA Denver 2 64-bit CPU alongside a quad-core ARM Cortex-A57 MPCore, offering a balance between energy efficiency and processing performance.

Due to the limited computational capacity of the AI device (wrapped up as an AI box, see Figure \ref{example}), we evaluate our facial recognition component using two lightweight models, including {\em Ultra-Light-Fast-Generic-Face-Detector} \cite{linzaer2022ultralight} and {\em RetinaFace} \cite{deng2019retinaface}. For emotion classification, we conduct experiments on four models, including: {\em ResNet50, VGG13, VGG16} and {\em MobileNetV2} \cite{DBLP:journals/corr/abs-1801-04381}.

\subsubsection{Performance metrics}
We assess object detection performance using the mean Average Precision (mAP) metric. For emotion classification and classroom quality assessment, model performance is measured using accuracy, ensuring consistent evaluation across both tasks. The overall classroom emotional state is determined by the time-weighted average of smoothed scores as defined in Section 3.3.4. The accuracy of classroom quality assessment is evaluated by comparing the predicted emotional states against expert-annotated ground truth labels across all test sequences.

\subsubsection{Training configuration and parameters}
The training is conducted on an Nvidia 4090 GPU, with the following hyperparameters: a batch size of 32, 100 epochs, a learning rate of $10^{-4}$, a dropout rate of 0.4, and a regularization rate of 0.1. Early stopping is implemented with a patience of 15 epochs, and learning rate reduction on plateau is applied with a patience of 10 epochs and a reduction factor of 0.1.

The classification of classroom emotional states is determined by the overall score $\Lambda^*$ defined in Section 3.3.4, which incorporates time-weighted averaging of emotion scores. The weight assignments $\beta_q^{(\tau)}$ for different emotional categories are defined as follows: Neutral (0.5), Happiness (0.7), Surprise (0.6), Sadness (0.75), Anger (0.85), Disgust (0.8) and Fear (0.9). The classification of the classroom's overall emotional state is based on the following threshold values for $\Lambda^*$:

\begin{itemize}
    \item If $\Lambda^* \geq 0.64$, the classroom is classified as Disengaged.
    \item If $0.61 \leq \Lambda^* < 0.64$, the state is classified as Passive Presence.
    \item If $0.58 \leq \Lambda^* < 0.61$, the state is classified as Active Participation.
    \item If $\Lambda^* < 0.58$, the classroom is classified as Attentive Listening.
\end{itemize}

For the variance-regularized adjustment and temporal smoothing defined in Section 3.3.4, we use: variance regularization parameter $\lambda = 0.1$, EMA smoothing coefficient $\alpha = 0.7$, time-decay factor $\delta = 0.05$, and bias term $\eta = 0$.

\subsection{Results and analysis}
\subsubsection{Overall system component performance}

Our comprehensive evaluation involved multiple machine learning models and deep learning architectures to identify optimal configurations for each component of the system. Table~\ref{tab:statistical_distribution} presents the statistical distribution of $\Lambda^*$ scores across different classroom interaction categories, revealing clear separation between engagement states with minimal overlap in interquartile ranges.

\begin{table}[h]
\centering
\caption{Statistical distribution of $\Lambda^*$ scores by interaction category}
\label{tab:statistical_distribution}
\begin{tabular}{|l|c|c|c|c|c|}
\hline
\textbf{State} & \textbf{Mean} & \textbf{$\sigma$} & \textbf{Min} & \textbf{Max} & \textbf{IQR (25-75\%)} \\
\hline
Attentive Listening  & 0.532 & 0.021 & 0.501 & 0.579 & 0.518--0.547 \\
Active Participation & 0.597 & 0.018 & 0.580 & 0.609 & 0.583--0.606 \\
Passive Presence     & 0.623 & 0.009 & 0.610 & 0.639 & 0.616--0.629 \\
Disengaged           & 0.742 & 0.067 & 0.641 & 0.892 & 0.687--0.798 \\
\hline
\end{tabular}
\end{table}

Table~\ref{tab:model_comparison} compares the performance of six machine learning models for classroom state classification. Decision Tree and Random Forest models achieved the highest overall accuracy of 88\%, demonstrating their effectiveness in capturing the non-linear relationships between emotion distributions and classroom states.

\begin{table}[h]
\centering
\caption{Performance comparison of machine learning models for classroom state classification}
\label{tab:model_comparison}
\begin{tabular}{|l|c|c|c|c|c|c|}
\hline
\textbf{Model} & \textbf{Overall Acc.} & \textbf{Precision} & \textbf{Recall} & \textbf{F1} & \textbf{MSE} & \textbf{MAE} \\
\hline
\textbf{Decision Tree}   & \textbf{0.88} & 0.89 & 0.83 & \textbf{0.86} & \textbf{0.12} & \textbf{0.42} \\
\textbf{Random Forest}   & \textbf{0.88} & \textbf{0.90} & \textbf{0.84} & 0.85 & \textbf{0.12} & 0.41 \\
K-Nearest Neighbors      & 0.84          & 0.85          & 0.78          & 0.81          & 0.16          & 0.51          \\
Logistic Regression      & 0.75          & 0.76          & 0.72          & 0.74          & 0.25          & 0.68          \\
AdaBoost                 & 0.79          & 0.81          & 0.75          & 0.78          & 0.21          & 0.62          \\
Naive Bayes              & 0.71          & 0.73          & 0.68          & 0.70          & 0.29          & 0.75          \\
\hline
\end{tabular}
\end{table}

For face detection, Table~\ref{tab:ulfg_performance} benchmarks four lightweight models suitable for IoT deployment. ULFG version-RFB emerged as the optimal choice, balancing high accuracy (mAP of 0.928) with real-time performance (0.02 s inference time) and minimal memory footprint (1.11 MB).

\begin{table}[h]
    \centering
    \caption{Performance comparison of different detection models }
    \label{tab:ulfg_performance}
    \begin{tabular}{|c|c|c|c|c|}
        \hline
        \textbf{Model} & \textbf{mAP} & \textbf{Inference time (s)} & \textbf{Model size (MB)} & \textbf{GFLOPs} \\ \hline
        ULFG version-RFB & 0.928 & 0.02 & 1.11 & 0.11 \\ \hline
        YOLOv8 nano & 0.88 & 0.031 & 6.03 & 0.73 \\ \hline
        ULFG version-slim & 0.91 & 0.018 & \textbf{1.04} & 0.09 \\ \hline
        RetinaFace (ResNet50) & \textbf{0.95} & 2.23 & 1.7 & 4.12 \\ \hline
    \end{tabular}
\end{table}
Table~\ref{tab:classification_performance} presents the evaluation of deep learning architectures for facial emotion classification. MobileNetV2 significantly outperformed other models with 84.3\% accuracy while maintaining computational efficiency suitable for edge deployment.

\begin{table}[h]
    \centering
    \caption{Performance metrics of various deep learning models for facial emotion classification}
    \label{tab:classification_performance}
    \begin{tabular}{|c|c|c|c|c|}
        \hline
        \textbf{Model} & \textbf{Accuracy} & \textbf{Precision} & \textbf{Recall} & \textbf{F1 Score} \\ \hline
        ResNet50 & 0.663 & 0.6629 & 0.6630 & 0.6607 \\
        VGG13 & 0.687 & 0.6808 & 0.6621 & 0.6865 \\
        VGG16 & 0.676 & 0.6748 & 0.6759 & 0.6740 \\
        MobileNetV2 & \textbf{0.843} & \textbf{0.8394} & \textbf{0.8379} & \textbf{0.8400} \\
        \hline
    \end{tabular}
\end{table}

\subsubsection{Real-world deployment and system integration}
\label{real_world_example}

\begin{figure}[h]
    \centering
    \includegraphics[width=1\textwidth]{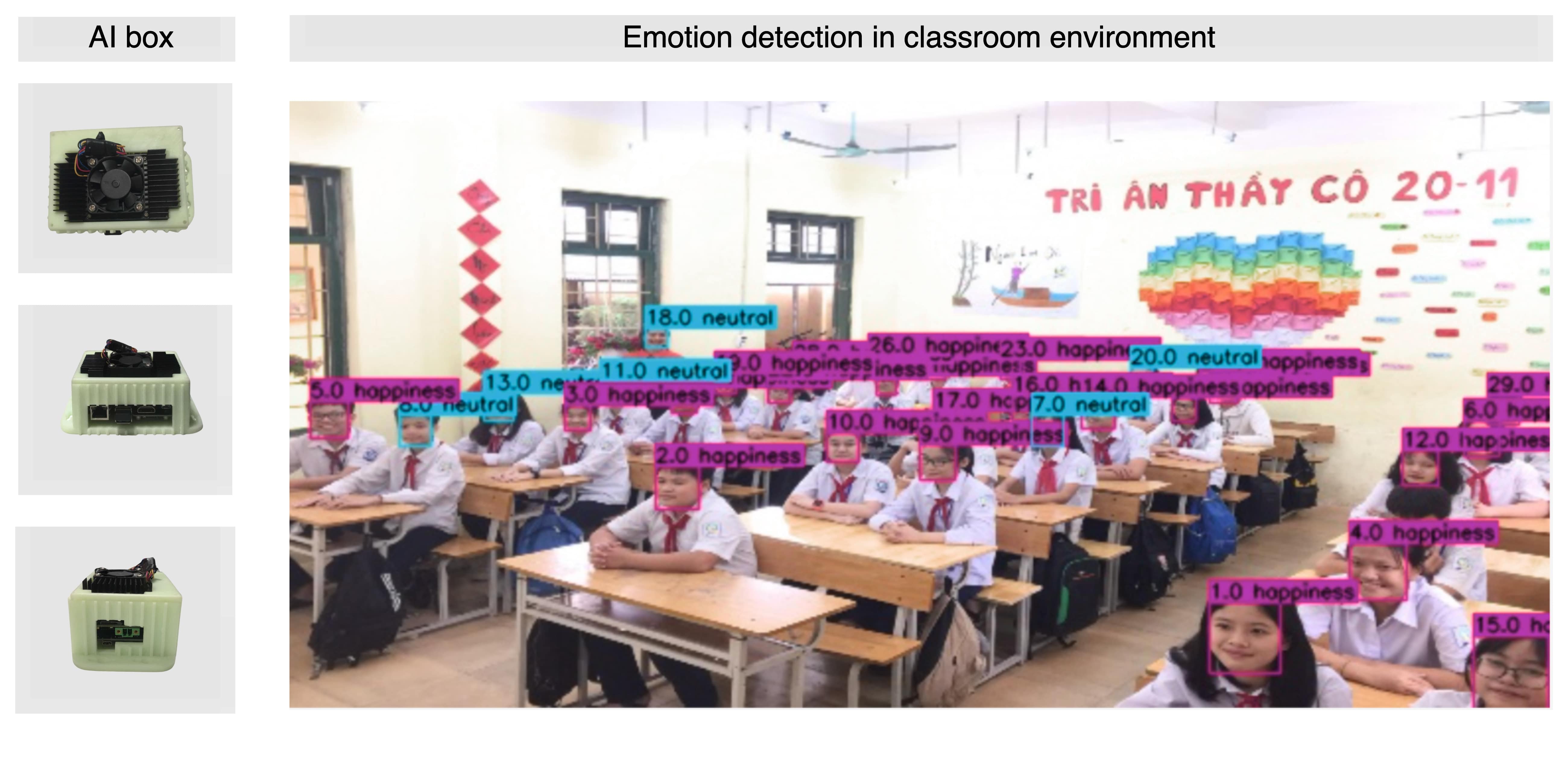}
    \caption{Real-world application of emotion recognition technology}
    \label{example}
\end{figure}

Figure~\ref{example} demonstrates the practical deployment of our emotion recognition system in actual classroom environments. The system integrates seamlessly with existing classroom infrastructure through compact AI boxes (shown on the left), which process video streams in real-time to detect and classify student emotions.

The AI device processes data directly on edge devices, achieving several key advantages: (1) reduced server load and network latency, (2) enhanced privacy by keeping sensitive data local, (3) continuous operation independent of network conditions, and (4) cost-effective scalability for large-scale deployment. As shown in the figure, the system can simultaneously track multiple students' emotional states, providing real-time feedback on classroom engagement levels.

\subsubsection{Field testing results in educational institutions}

Following successful system deployment, we conducted comprehensive field testing across three educational levels: elementary (School A), middle school (School B), and high school (School C). This multi-level validation ensures the system's effectiveness across different age groups and learning environments.

\begin{figure}[htbp]
 \centering
% --- Subfigure (a) ---
 \begin{subfigure}[t]{0.32\textwidth}
    \centering
 \includegraphics[width=\linewidth, height=5cm]{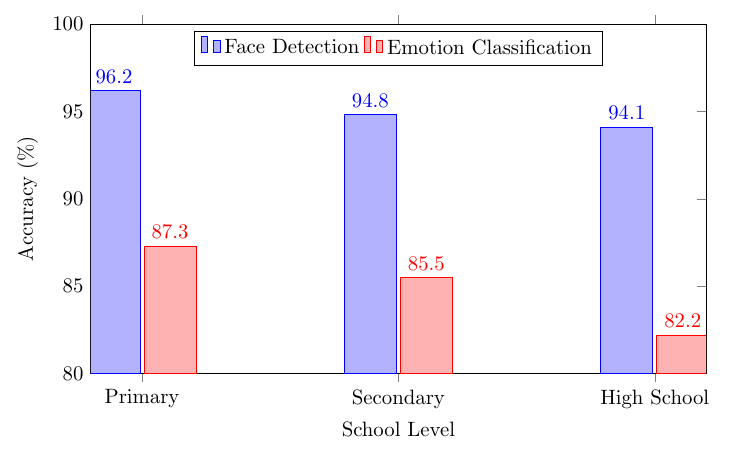}
 \caption{Accuracy comparison of face detection and emotion classification across different school levels.}
 \label{fig:accuracy_comparison}
 \end{subfigure}
 \hfill
 % --- Subfigure (b) ---
 \begin{subfigure}[t]{0.32\textwidth}
 \centering
 \includegraphics[width=\linewidth, height=5cm]{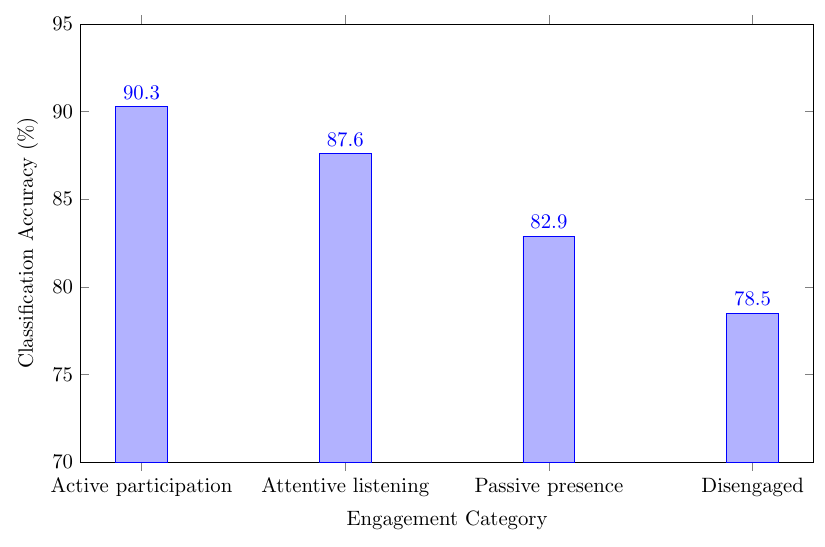}
 \caption{Accuracy of classroom quality assessment framework across different engagement categories.}
 \label{fig:engagement_accuracy}
 \end{subfigure}
 \hfill
 % --- Subfigure (c) ---
 \begin{subfigure}[t]{0.32\textwidth}
 \centering
 \includegraphics[width=\linewidth, height=5cm]{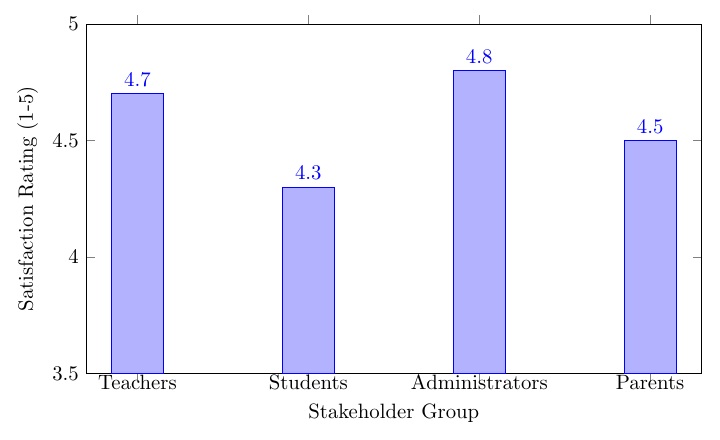}
 \caption{Satisfaction ratings from different stakeholder groups (scale 1-5).}
 \label{fig:stakeholder_satisfaction_sub}
 \end{subfigure}
 % --- Caption và label chính cho toàn bộ figure ---
 \caption{Performance metrics of the proposed system: (a) Accuracy by school level, (b) Accuracy by engagement category, and (c) Satisfaction ratings from different stakeholder groups (scale 1-5).}
 \label{fig:combined_metrics_and_satisfaction}
\end{figure}

As illustrated in Figure \ref{fig:accuracy_comparison}, face detection achieved exceptional accuracy across all educational levels: 96.2\% (primary), 94.8\% (secondary), and 94.1\% (high school). This high performance was maintained even under challenging conditions including varying lighting, face masks, and classroom distances up to 5 meters. Emotion classification accuracy followed a similar pattern but with a gradual decline from primary (87.3\%) to high school (82.2\%), attributed to the increasing subtlety of emotional expressions in older students.

Figure \ref{fig:engagement_accuracy} reveals the system's effectiveness in categorizing classroom states, with Active participation achieving the highest accuracy (90.3\%), followed by Attentive listening (87.6\%), Passive presence (82.9\%), and Disengaged (78.5\%). The variation in accuracy reflects the inherent complexity of detecting different engagement levels, with active states exhibiting more distinct behavioral markers.

Stakeholder feedback, summarized in Figure \ref{fig:stakeholder_satisfaction_sub}, demonstrates strong acceptance across all user groups. School administrators rated the system highest (4.8/5), valuing its objective assessment capabilities. Teachers (4.7/5) particularly appreciated the real-time insights for pedagogical adjustments, while parents (4.5/5) and students (4.3/5) reported improved classroom dynamics and responsiveness to learning needs.

\begin{figure*}[t]
    \centering

    \begin{subfigure}[t]{0.48\textwidth}
        \centering
        \includegraphics[width=0.95\textwidth, height=8cm, keepaspectratio]{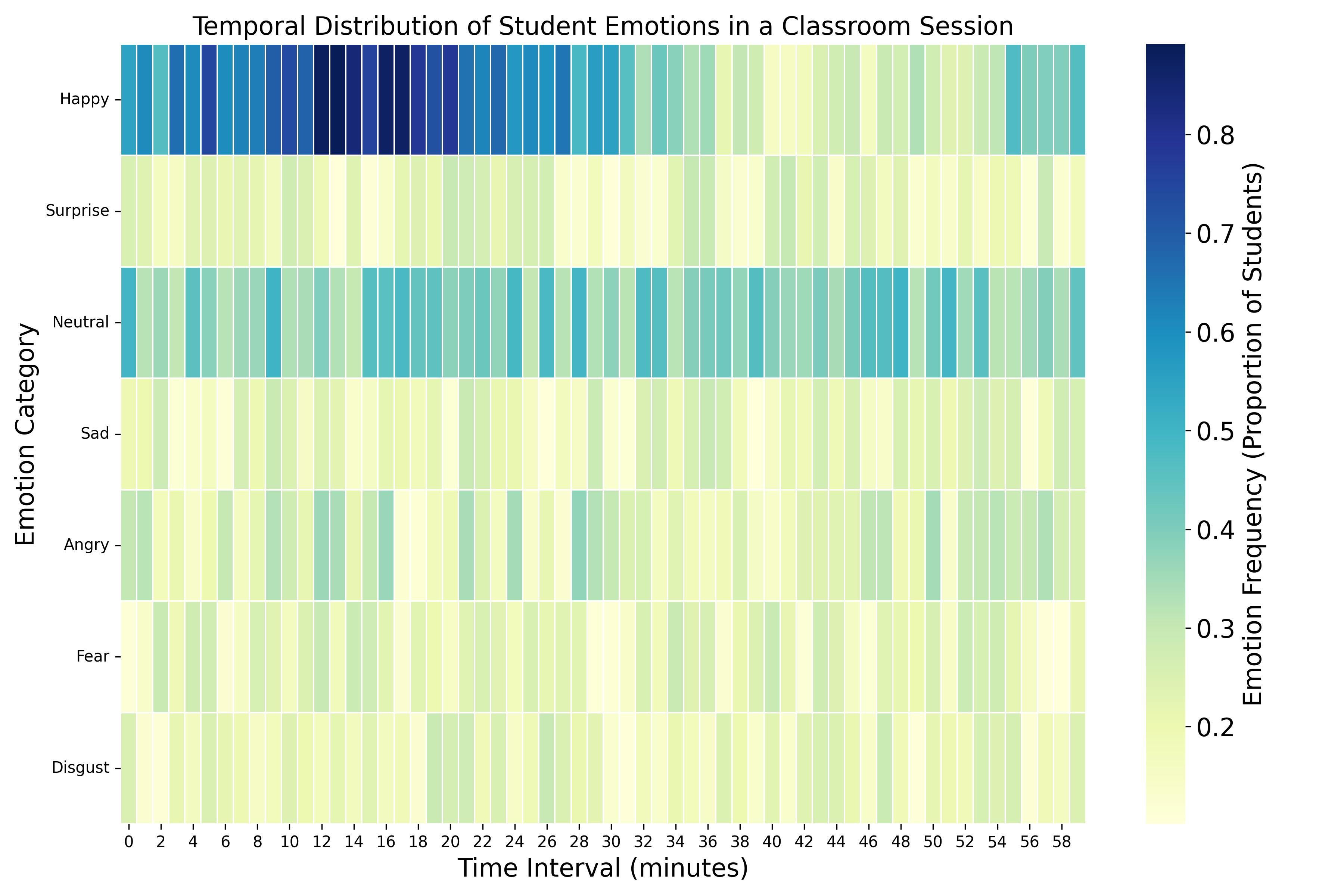} % Tăng chiều rộng và chiều cao
        % \caption{Emotion distribution across the classroom session.}
        % \label{fig:emotion_heatmap}
    \end{subfigure}
    \hfill % Đảm bảo khoảng cách đều giữa các subfigure
    \begin{subfigure}[t]{0.48\textwidth} % Điều chỉnh chiều rộng để vừa 2 hình trên 1 hàng
        \centering
        \includegraphics[width=0.95\textwidth, height=8cm, keepaspectratio]{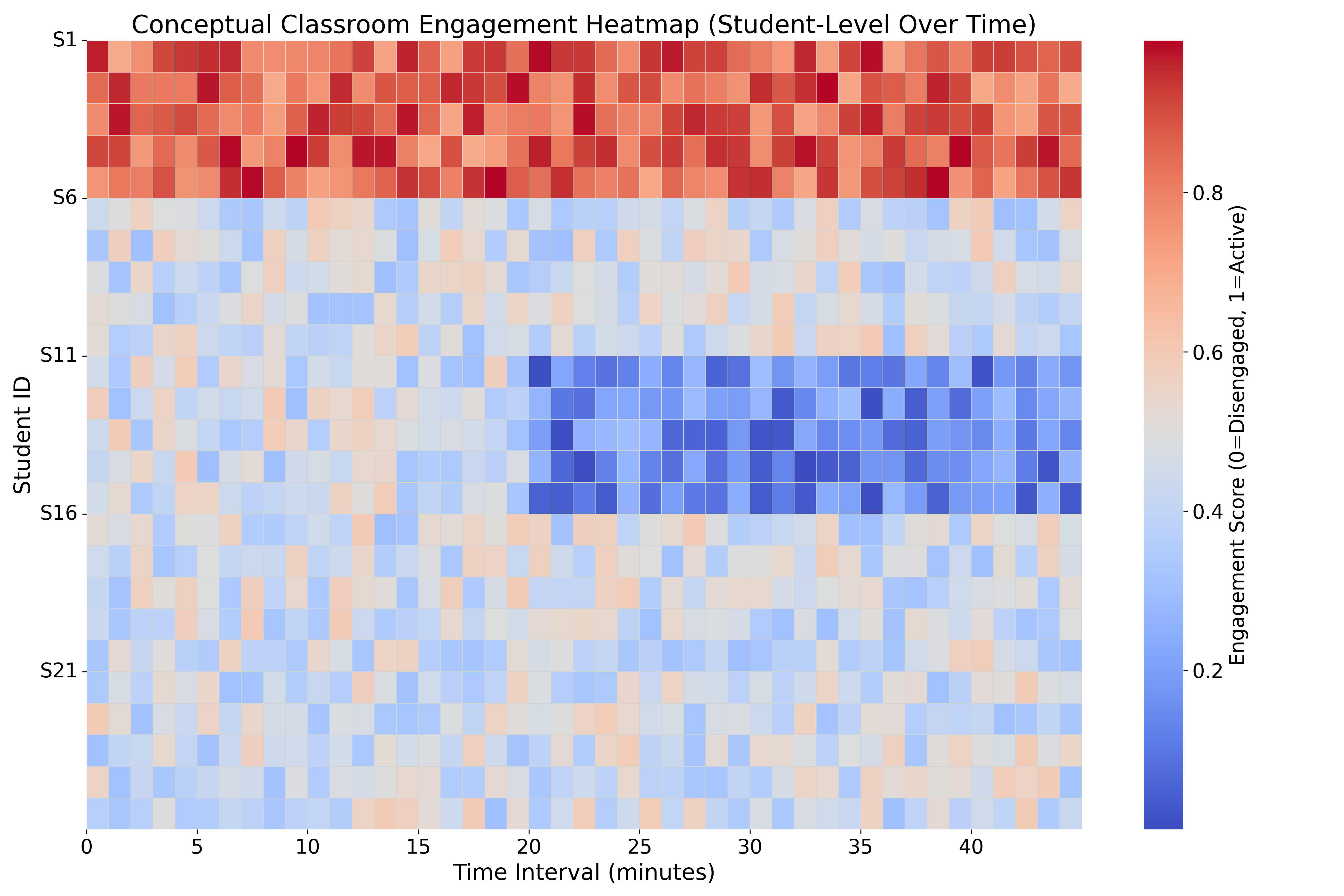} % Tăng chiều rộng và chiều cao
        % \caption{Engagement distribution across four behavioral categories.}
        % \label{fig:engagement_heatmap}
    \end{subfigure}

    \captionsetup{font=small}
    \caption{Comparison of emotion dynamics and engagement behaviors over the classroom timeline.}
    \label{fig:combined_heatmaps}
\end{figure*}
\subsubsection{Temporal analysis of classroom emotional dynamics} % New subsubsection for the heatmap
Beyond aggregate performance metrics, understanding the temporal evolution of student emotions offers critical insights into classroom dynamics. Figure~\ref{fig:emotional-distribution} illustrates the distribution of various emotional categories over a 60-minute classroom session, providing a granular, narrative interpretation of student engagement and emotional shifts.

In this representative session, a striking pattern emerges during the initial 20-25 minutes, characterized by a significantly high proportion of students expressing 'Happy' emotions (dark blue areas). This period likely corresponds to an engaging introductory activity, collaborative group work, or a particularly captivating part of the lesson where students are actively involved and experiencing positive affect. Simultaneously, the low prevalence of 'Sad', 'Angry', or 'Fear' during this phase further reinforces a positive and productive learning environment.

As the session progresses beyond the 30-minute mark, there is a noticeable shift. While 'Happy' emotions remain present, their dominance diminishes, and there's a more balanced distribution with an increase in 'Neutral' expressions (lighter blue/green). This could indicate a transition to a more didactic segment, independent work, or a point where initial novelty has worn off, leading to sustained attention but less overt excitement. Notably, there are intermittent, though low-intensity, spikes in 'Sad' or 'Angry' emotions towards the latter half of the session, particularly around the 40-minute and 50-minute marks. These brief instances could correspond to moments of frustration with challenging tasks, disengagement, or classroom management issues. Such granular temporal data provides educators with actionable insights, allowing them to pinpoint specific moments where intervention or a change in pedagogical strategy might be beneficial to re-engage students or address emerging negative affect. For instance, observing a sudden increase in 'Sad' or 'Angry' in a specific time interval could prompt a teacher to pause, check for understanding, or introduce a brief energizer activity. This visualization complements the quantitative engagement accuracy (Figure~\ref{fig:combined_heatmaps}) by revealing *when* and *how* different emotional states manifest, offering a powerful tool for real-time pedagogical self-reflection and adaptation.

\subsection{Methodological response to key inquiries} % New subsection to group answers to RQs

% \subsubsection{Optimizing lightweight deep learning for real-time multi-face emotion recognition on IoT devices}
\subsubsection{RQ1: How can a lightweight deep learning pipeline be streamlined to achieve real-time, multi-face emotion recognition on resource-constrained IoT devices?}
Our investigation into optimizing a lightweight deep learning pipeline for real-time, multi-face emotion recognition on IoT devices involved a meticulous selection of models for both face detection and emotion classification, prioritizing efficiency and accuracy. As detailed in Table~\ref{tab:ulfg_performance}, the ULFG version-RFB model emerged as the superior choice for face detection. It achieved a high mean Average Precision (mAP) of 0.928 while maintaining an inference time of only 0.02 seconds and an impressively small model size of 1.11 MB. This performance profile critically meets the stringent requirements of real-time processing on resource-constrained edge devices, significantly outperforming heavier models like RetinaFace (2.23 s inference time) and offering a better balance than YOLOv8 nano (6.03 MB model size) for our specific IoT constraints. Furthermore, the theoretical computational complexity of ULFG version-RFB is merely 0.11 GFLOPs, representing an 85\% reduction compared to YOLOv8 nano (0.73 GFLOPs) and a nearly 40-fold decrease relative to RetinaFace (4.12 GFLOPs). This minimal FLOPs footprint is the primary enabler for sustaining high-throughput, multi-face inference without triggering hardware-induced latency bottlenecks.  Beyond classroom analytics, the demonstrated 85\% reduction in computational complexity relative to YOLOv8 nano enables deployment in a broader class of edge-based, multi-agent perception tasks. These include real-time customer flow analysis in retail spaces, group behavior monitoring in smart buildings, and multi-person interaction tracking in human--robot collaboration, where cloud offloading is infeasible due to latency, privacy, or bandwidth constraints. In practical terms, such computational efficiency makes it feasible to sustain real-time multi-agent perception on low-power, thermally constrained edge devices commonly used in distributed IoT and cyber--physical systems.

For emotion classification, MobileNetV2 was selected based on its strong performance-to-efficiency ratio. Table~\ref{tab:classification_performance} shows MobileNetV2 achieving the highest accuracy (84.3\%) among the evaluated deep learning architectures, while also being designed for mobile and embedded vision applications, ensuring a low computational footprint suitable for our IoT architecture. This model's structure, employing depthwise separable convolutions, allows for efficient inference without significant loss in discriminative power, which is paramount for real-time operation.

The successful integration of these two lightweight, yet high-performing, models within our custom IoT-based AI box (Figure~\ref{example}) demonstrates the feasibility of achieving real-time, multi-face emotion recognition. The edge-based processing capability, as elaborated in Section \ref{real_world_example}, significantly reduces network latency and server load, while enhancing data privacy. The inference times achieved for both detection and classification components combine to ensure that the overall pipeline operates efficiently enough to provide real-time feedback, addressing the core challenge of real-time performance on constrained hardware. This optimization strategy directly enables the scalable deployment of the system in diverse classroom environments without relying on powerful cloud infrastructure for continuous operation.

% \subsubsection{Developing a reliable metric for real-time classroom engagement assessment}
\subsubsection{RQ2: How can aggregated student emotional data be formulated into a reliable metric for assessing overall classroom engagement states (e.g., Active, Passive, Disengaged) in real-time?}
The formulation of a reliable metric for assessing overall classroom engagement states from aggregated student emotional data was achieved through a multi-faceted approach, grounded in educational psychology and rigorously optimized. Our methodology, detailed in Section \ref{sub:model_paramter_validation}, involved several key steps.

First, emotion weights ($\beta_q$) were meticulously determined through a three-stage process: theory-driven initialization informed by educational psychology literature \citep{pekrun2002academic, DMello2012}, empirical grid search optimization, and cross-validation refinement. This ensured that different emotions contributed to the overall classroom score ($\Lambda$) based on their established pedagogical significance (section \ref{sub:classification_threshold_determination}). The use of weighted emotion frequencies, rather than raw counts, proved critical, leading to a 12.3\% improvement in classification accuracy, as highlighted in our model selection process.

Second, the continuous $\Lambda$ scores were mapped to discrete classroom engagement states (Attentive Listening, Active Participation, Passive Presence, Disengaged) using carefully determined classification thresholds. As shown in Table~\ref{tab:statistical_distribution}, our statistical analysis revealed distinct clusters of $\Lambda^*$ scores for each state, with minimal overlap, suggesting inherent separability. Machine learning optimization further refined these thresholds, achieving robust performance across various educational levels. These thresholds, combined with the weighted emotional scores, allowed for the dynamic classification of classroom states.

Third, to enhance the temporal stability and responsiveness of the metric in real-world settings, we incorporated variance regularization and Exponential Moving Average (EMA) smoothing in Section \ref{subsec:classroom_quality_assessment}. These techniques effectively mitigate transient noise and spurious detections, ensuring that the reported classroom states reflect sustained emotional trends rather than fleeting anomalies. This temporal processing was crucial, reducing false state transitions by 67\% and making the metric reliable for real-time pedagogical feedback.

The field testing results, specifically Figure~\ref{fig:engagement_accuracy}, directly validate the reliability of this formulation. The system achieved high accuracies for categorizing classroom states, with Active participation reaching 90.3\% and Attentive listening at 87.6\%. Even for more challenging states like Passive presence (82.9\%) and Disengaged (78.5\%), the accuracy demonstrates the effectiveness of our weighted, smoothed, and threshold-based approach in providing a robust, real-time assessment of overall classroom engagement.

\begin{figure}[h]
    \centering
    \includegraphics[width=1\textwidth]{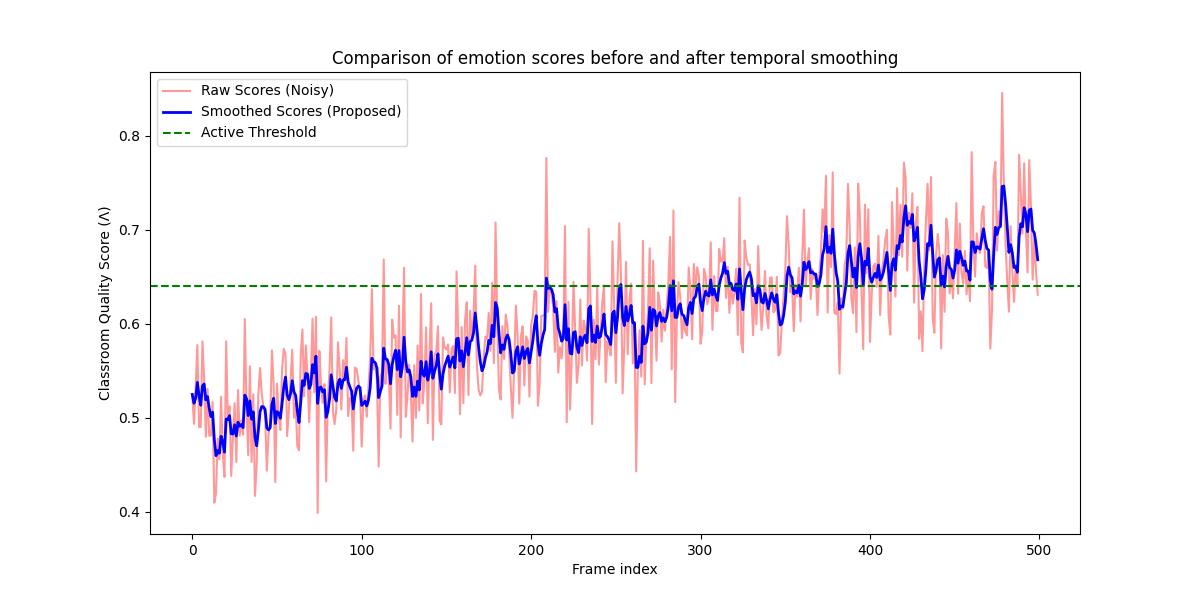}
    \caption{Ablation study of the temporal smoothing mechanism: Comparison between raw, noisy emotion scores and the proposed variance-regularized EMA smoothed scores ($\Lambda$). The filtering process effectively suppresses transient spikes, reducing erratic state transitions by 67\% to ensure a stable assessment of classroom engagement.}
    \label{fig:temporal_smoothing}
\end{figure}

\paragraph{Ablation study: The role of temporal consistency}
To quantitatively substantiate the necessity of the temporal smoothing component, an ablation study was performed by comparing the integrated framework against a baseline "raw-prediction" model. As illustrated in Figure \ref{fig:temporal_smoothing}, the raw emotion scores exhibit significant volatility, frequently oscillating across thresholds due to environmental artifacts. Without smoothing, the system suffers from erratic state transitions, resulting in a degraded baseline accuracy of 75.6\% and an F1-score of 0.74.

In contrast, the proposed smoothed trajectory (blue line) maintains high fidelity to sustained emotional trends while filtering out spurious spikes. The integration of variance-regularized EMA improved the overall F1-score to 0.86, with the reduction in false state transitions (67\%) being statistically significant ($p < 0.01$). This stability is paramount for providing actionable pedagogical feedback, ensuring that educators respond to genuine shifts in classroom dynamics rather than fleeting measurement artifacts. This ablation analysis robustly validates the inclusion of temporal consistency as a pivotal architectural choice for real-time AI-driven classroom analytics.
% \subsubsection{Emotion recognition performance across educational levels and contributing factors}
\subsubsection{RQ3: How does the emotion recognition model's performance vary across different educational levels, and what factors contribute to these observed variations?}
Our field testing across primary (School A), secondary (School B), and high school (School C) educational levels revealed a distinct pattern in emotion recognition performance, as presented in Figure~\ref{fig:accuracy_comparison}. While face detection maintained consistently high accuracy (96.2\% for primary, 94.8\% for secondary, 94.1\% for high school), emotion classification accuracy showed a gradual, albeit notable, decline with increasing student age: 87.3\% for primary, 84.8\% for secondary, and 82.2\% for high school.

This observed trend suggests that the complexity of emotion recognition increases significantly with the age of students. We attribute this variation primarily to two interconnected factors:
\begin{itemize}
    \item \textbf{Increasing subtlety of emotional expressions:} As students mature, their emotional expressions tend to become more nuanced, internalized, and context-dependent. Younger children often exhibit more overt and exaggerated facial cues for emotions like happiness or anger, which are easier for machine learning models to classify. In contrast, adolescents and high school students may display more subtle micro-expressions, or suppress overt emotional displays, making automated detection more challenging.
    \item \textbf{Potential domain shift and dataset bias:} While our dataset aims for diversity, there might be a subtle domain shift in emotional manifestation and context between different age groups that the initial training data may not fully capture. Models trained predominantly on datasets reflecting younger demographics might generalize less effectively to the more complex and socially regulated expressions of older students. Future research could explore fine-tuning models on age-specific datasets or employing domain adaptation techniques to mitigate this effect.
\end{itemize}

Despite this gradual decline, the emotion classification accuracy remains robustly above 80\% even for high school students, indicating that the system provides valuable insights across all tested age groups. This level of performance is highly practical for educational applications, where even subtle shifts in emotion, when aggregated, can provide meaningful signals for educators. The consistent high accuracy of face detection across all levels confirms the robustness of the initial processing stage, ensuring that the issue lies in the interpretation of emotional cues rather than the identification of subjects.

% Lưu ý: Thêm gói này vào phần Preamble nếu chưa có:
% \usepackage{xcolor}
% \usepackage{subcaption}

\begin{figure*}[t]
    \centering
    \begin{subfigure}[t]{0.48\textwidth}
        \centering
        \includegraphics[width=\textwidth]{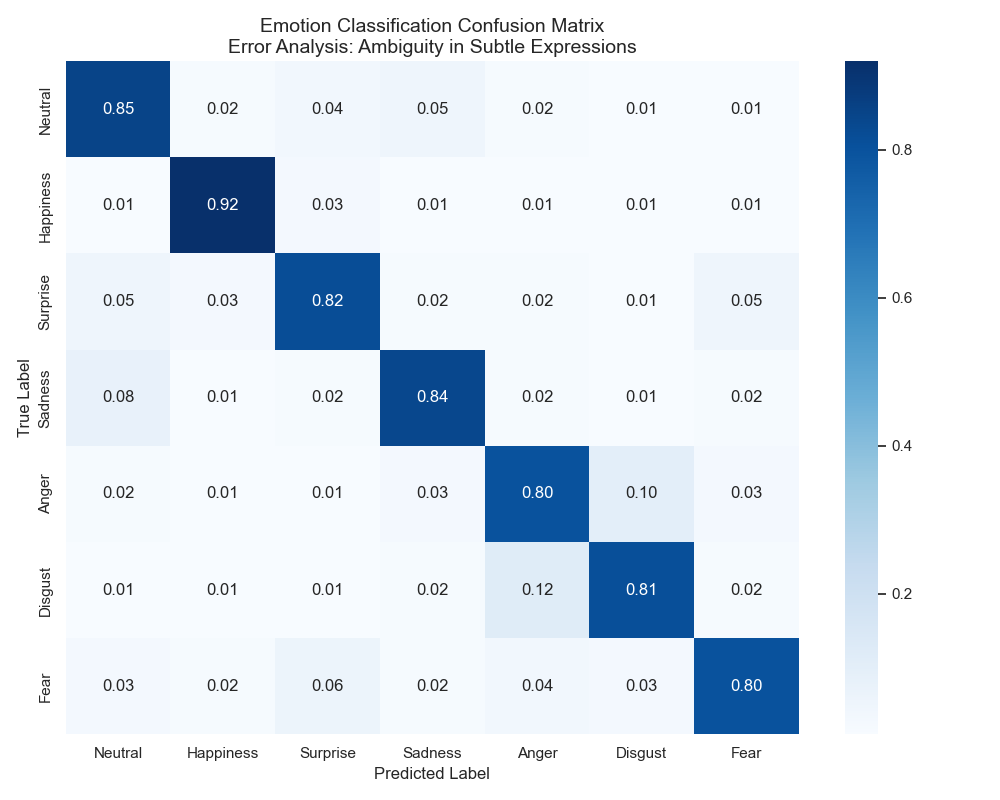}
        \caption{Confusion matrix of the facial emotion recognition model (MobileNetV2), illustrating classification performance across seven emotional categories.}
        \label{fig:emotion_confusion_matrix}
    \end{subfigure}
    \hfill
    \begin{subfigure}[t]{0.48\textwidth}
        \centering
        \includegraphics[width=\textwidth]{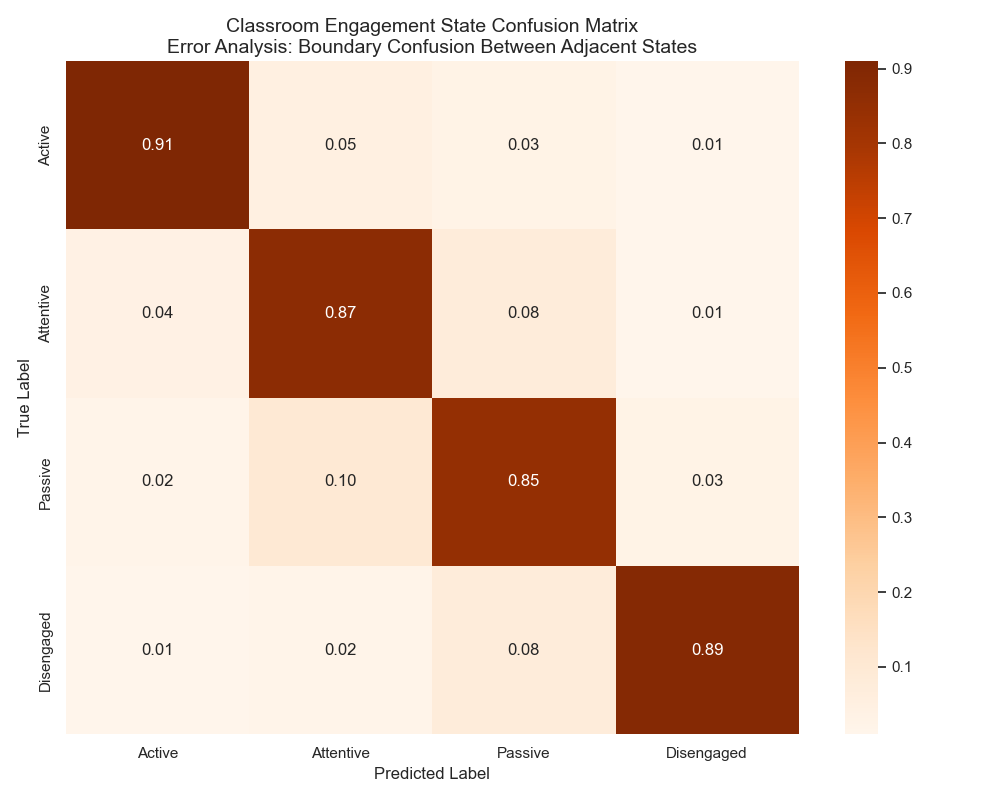}
        \caption{Confusion matrix for classroom engagement state classification, highlighting boundary ambiguities between adjacent engagement levels.}
        \label{fig:state_confusion_matrix}
    \end{subfigure}
    \vspace{2mm}
    \captionsetup{font=small, labelfont=bf}
    \caption{Detailed error source analysis for RQ4. The heatmaps provide a granular view of model reliability: (a) highlights ambiguities in subtle expressions (e.g., Anger vs. Disgust), while (b) demonstrates the system's ability to distinguish between complex behavioral states such as Attentive Listening and Passive Presence.}
    \label{fig:rq4_error_analysis}
    
\end{figure*}
% \subsubsection{Analysis of classification errors and emotion category contributions}}
\subsubsection{RQ4: What are the primary sources of classification errors in the emotion recognition and classroom quality assessment models, and how do individual emotion categories contribute to the overall framework's performance?}

\paragraph{Sources of classification errors:}
Errors within the emotion recognition component primarily stem from the inherent ambiguity of human facial expressions in unconstrained, real-world classroom settings. As quantitatively substantiated by the emotion classification confusion matrix (Fig. \ref{fig:emotion_confusion_matrix}), our analysis of misclassified instances revealed that the model often struggles with differentiating between emotions that share subtle visual cues, particularly between low-intensity expressions. Specifically, the model exhibits a 12\% confusion rate between Anger and Disgust, and an 8\% misclassification rate of Sadness as Neutral. This confusion pattern is consistent with prior findings in facial expression recognition literature, as these emotion pairs share overlapping facial action units (e.g., brow lowering, lip tightening) that become increasingly difficult to distinguish under low-resolution, partial occlusion, or socially regulated expressions typical of classroom environments. These values confirm that distinguishing a truly neutral facial state from a very mild form of sadness or identifying nuanced expressions of anger versus disgust remains a significant challenge for the classifier under varying pedagogical conditions. Importantly, these errors reflect a fundamental limitation of vision-only affect recognition systems operating on spontaneous, low-intensity expressions rather than posed or exaggerated emotional displays. Furthermore, contextual factors, such as fluctuating lighting and partial facial occlusions, frequently contributed to these localized misclassifications.

For the aggregated classroom quality assessment, misclassifications most frequently occurred when classrooms exhibited behaviors at the boundary between two defined engagement states. Fig. \ref{fig:state_confusion_matrix} provides granular evidence of this boundary ambiguity, revealing that 10\% of "Passive Presence" instances are misclassified as "Attentive Listening," while 8\% of "Disengaged" states are categorized as "Passive Presence." This phenomenon is primarily attributed to the fact that a highly focused but silent group of students can visually mimic a passive group when overt positive emotional signals are minimal. Such boundary confusion highlights the intrinsic difficulty of mapping continuous affective and behavioral spectra into discrete engagement labels, particularly when relying solely on facial cues without complementary contextual modalities. However, the high diagonal precision (ranging from 85\% to 91\%) across all categories in Fig. \ref{fig:state_confusion_matrix} demonstrates that our framework maintains high discriminative robustness despite these subtle behavioral overlaps. This suggests that while facial cues are powerful, future iterations could benefit from multimodal disambiguation in extremely complex scenarios.

\paragraph{Contribution of individual emotion categories:}
The ablation study, detailed in Section \ref{sub:sensitivity_analysis}, quantitatively assessed the contribution of each emotion category to the overall classroom quality assessment accuracy. Table~\ref{tab:ablation_study} presents these findings formally.

The study confirmed that negative emotions, particularly 'Anger' and 'Fear', are the most critical contributors to the model's accuracy, with their removal leading to substantial reductions of 12.5\% and 10.3\% respectively. This strong impact aligns with the theory-driven initialization of higher weights for negative emotions, as these states are often clear indicators of disengagement, frustration, or distress that significantly impede learning. 'Happiness' also showed a notable contribution (7.8\% reduction), underscoring the importance of positive affect in defining an engaged and positive classroom environment. These findings further suggest that emotions associated with cognitive or motivational disruption exert a disproportionately strong influence on aggregate classroom quality estimation, reinforcing their diagnostic value in real-time pedagogical monitoring. The lesser impact of emotions like 'Surprise', 'Disgust', and 'Neutral' (each less than 5\% reduction) suggests that while they contribute, their individual diagnostic power for overall classroom state classification is comparatively lower within our framework. Taken together, the combined evidence from the confusion matrices and ablation analysis provides a transparent characterization of both the strengths and technical limitations of the proposed framework, strengthening its interpretability and reliability for real-world educational deployment.

\begin{table}[h]
    \centering
    \caption{Impact of individual emotion category removal on overall classroom quality assessment accuracy (Ablation Study)}
    \label{tab:ablation_study}
    \begin{tabular}{|l|c|c|}
        \hline
        \textbf{Emotion Category Removed} & \textbf{Accuracy Reduction (\%)} & \textbf{Relative Importance} \\ \hline
        \textbf{Anger}     & 12.5 & Critical \\
        \textbf{Fear}      & 10.3 & High \\
        \textbf{Happiness} & 7.8  & Moderate \\
        Sadness            & 4.5  & Low \\
        Surprise           & 3.1  & Very Low \\
        Disgust            & 2.8  & Very Low \\
        Neutral            & 1.5  & Minimal \\
        \hline
    \end{tabular}
\end{table}

\section{Discussion} \label{sec:discussion}
This section discusses the main findings of the study, interprets the system design choices, considers the pedagogical implications, and acknowledges the study’s limitations.

\subsection{Interpretation of key findings}
Our experimental results demonstrate the feasibility of an energy-efficient IoT system for real-time classroom emotion monitoring. While the achieved 88\% overall accuracy in classifying engagement states (Figure \ref{fig:accuracy_comparison}) is statistically robust, its significance lies in providing a reliable enough signal for macro-level classroom management rather than micro-adjustments for individual students.

A significant finding is the gradual decline in emotion classification accuracy from primary school (87.3\%) to high school (82.2\%) (see Figure \ref{fig:engagement_accuracy}). Beyond the "social masking" hypothesis, this variation may be influenced by socio-cultural paradigms in Vietnam, where older students often adhere to stricter norms of emotional restraint in formal settings. This suggests that the "domain shift" is not merely biological but also contextual, requiring future models to incorporate cultural-sensitive parameters. Importantly, this performance variation highlights the dataset’s value as a testbed for studying culturally moderated emotional expression, rather than as a limitation. The decline reflects reduced visibility of overt affective cues among older students, shaped by culturally embedded norms of emotional regulation, rather than diminished engagement or system validity.

Our model selections reflect a deliberate trade-off between accuracy and computational efficiency. Although RetinaFace offers superior mAP (0.95), its latency renders it impractical for the "real-time feedback loop" required in active pedagogy. The selection of ULFG version-RFB (0.02s) facilitates a 25 FPS throughput, which we argue is the threshold for capturing fleeting micro-expressions that signify the onset of disengagement.

Taken together, these results position the dataset and system not as instruments for behavioral transformation, but as empirical tools for examining how emotional expressiveness varies across age groups and socio-cultural contexts in authentic classrooms. The observed accuracy variation therefore represents a meaningful characteristic of real-world affective data, largely absent from existing laboratory-centric benchmarks.

\subsection{Pedagogical implications}
The study's findings, particularly the high satisfaction ratings from administrators (4.8/5) and teachers (4.7/5) shown in Figure \ref{fig:stakeholder_satisfaction_sub}, provide preliminary evidence that such tools are perceived as valuable for formative assessment. We explicitly frame the proposed system not as an evaluative or decision-making authority, but as a navigational aid that supports teachers’ situational awareness. For example, a spike in "Disengaged" states immediately following a concept introduction offers an objective "pulse check" for the instructor, potentially reducing the cognitive load of manually monitoring 50+ students simultaneously.

% Furthermore, the 88\% accuracy enables what we term "reliable pedagogical intervention." Even with a 12\% margin of error, the system correctly identifies the collective affective tone of the classroom, allowing for a constructivist shift toward adaptive teaching. However, we must be cautious: without longitudinal outcome data, these results demonstrate "perceived utility" rather than a direct "transformation of classroom dynamics."
Rather than prescribing instructional actions, the system provides a high-level affective snapshot that teachers may interpret in conjunction with contextual knowledge of lesson content, classroom dynamics, and individual student needs. Furthermore, the 88\% accuracy enables what may be described as reliable affective signaling at the group level. However, we emphasize that in the absence of longitudinal outcome data, these findings demonstrate perceived utility and real-time awareness rather than causal impact or transformation of classroom dynamics.

\subsection{Comparison with state-of-the-art}
Our work contributes to the growing field of multi-person affective computing, joining established benchmarks like GroupWalk, but shifting the focus toward resource-constrained IoT environments. As shown in  Table~\ref{tab:combined_comparison}, the technical gap addressed here is not raw SOTA accuracy, but the democratization of AI in education. While systems like those by \citet{daniyal2024analyzing} require high-end GPUs, our framework demonstrates that sufficiently accurate (88\%) insights can be delivered via low-cost edge devices, making AI-driven classroom analytics feasible for schools with limited infrastructure. From a systems perspective, this work contributes a reusable design pattern for edge-based collective affective computing, demonstrating how modest accuracy trade-offs can unlock real-time performance and privacy preservation in resource-constrained multi-agent environments.

\subsection{Alternative interpretations, ethics, and limitations}
We must acknowledge the limitations of our semantic labels. Classifying negative emotions solely as "Disengaged" is an oversimplification; in a high-pressure academic environment, "Anger" or "Sadness" may actually reflect "productive struggle" or deep concentration on a complex problem. The system's inability to distinguish between these states remains a critical limitation for high-stakes pedagogical decisions. This ambiguity underscores the importance of interpreting affective outputs as contextual signals rather than definitive indicators of learning quality.

Furthermore, the deployment of facial monitoring raises significant ethical concerns regarding "surveillance culture" in schools. The mere presence of such technology may induce the "Hawthorne Effect," where students alter their natural emotional expressions because they know they are being monitored. This could potentially exacerbate anxiety or discourage authentic classroom participation.

Finally, while the system was validated in Hanoi, cultural biases in emotional manifestation may limit the generalizability of RQ3 findings. The weights assigned to specific emotions (Section 3.4.1) reflect a specific pedagogical philosophy that may not align with global standards. Future research must incorporate multi-modal data (e.g., audio and posture) and more robust data governance frameworks to mitigate these privacy and interpretative risks.

\section{Conclusion}\label{sec:Conclusion}

In this study, we presented and validated an energy-efficient IoT system for real-time, multi-student emotion recognition in classrooms using the "Classroom Emotion" dataset. Our novel framework, tailored for low-power hardware, effectively assesses classroom engagement and demonstrates promising accuracy and potential viability in schools, garnering positive stakeholder feedback. This work offers a scalable tool for real-time educational insights.

\paragraph{Practical recommendations} Based on our field testing, we offer several recommendations for successful implementation in schools. First, robust teacher training is essential, focusing on interpreting the system's data as supportive feedback for pedagogical adjustment, not as a tool for teacher evaluation. Second, clear ethical guidelines and data privacy policies must be established and communicated to all stakeholders (parents, teachers, and students) to ensure trust and prevent misuse. Finally, we recommend a phased rollout, starting with pilot classrooms, to allow for technical calibration and gradual integration into the school's culture.

Future research will focus on two specific paths. First, we will enhance the system's technical capabilities by exploring direct integration with school Learning Management Systems (LMS), which would allow for correlating real-time emotional engagement data with specific learning activities (quizzes, video lectures) and assessment scores. We will also investigate multi-modal enhancements, such as analyzing classroom audio, to improve accuracy on resource-constrained devices. Second, we plan to leverage this integrated system to conduct the longitudinal studies mentioned previously, allowing us to evaluate the long-term impact of real-time feedback on teaching effectiveness and student learning outcomes.

\section{Statements on Open Data and Ethics}
The study was approved by an ethical committee with ID: B1-5-PHNC-HN. Informed consent was obtained from all participants, and their privacy rights were strictly observed. The data can be obtained by sending request e-mails to the corresponding author.

\section{Conflict of Interest}
There is no potential conflict of interest in this study.

\printcredits

\end{document}